\pdfoutput=1

\documentclass[11pt]{article}

\usepackage{acl}

\usepackage{times}
\usepackage{latexsym}

\usepackage[T1]{fontenc}

\usepackage[utf8]{inputenc}

\usepackage{microtype}

\usepackage{inconsolata}

\usepackage{comment}
\usepackage{graphicx}
\usepackage{amsmath}
\usepackage{amsfonts}

\usepackage{caption}
\usepackage{subcaption}

\begin{document}

%
%

\title{Style Vectors for Steering Generative Large Language Models}

\author{Kai Konen \quad  Sophie Jentzsch \quad  Diaoulé Diallo \quad  Peer Schütt \\ \bf Oliver Bensch \quad Roxanne El Baff \quad Dominik Opitz \quad Tobias Hecking \\
        Institute for Software Technology, German Aerospace Center (DLR) \\ 
        \texttt{\{first\}.\{last\}@dlr.de} \\
        } 

\maketitle

\begin{abstract}
This research explores strategies for \textit{steering} the output of large language models (LLMs) towards specific styles, such as sentiment, emotion, or writing style, by adding \textit{style vectors} to the activations of hidden layers during text generation. We show that style vectors can be simply computed from recorded layer activations for input texts in a specific style in contrast to more complex training-based approaches. 
Through a series of experiments, we demonstrate the effectiveness of \textit{activation engineering} using such \textit{style vectors} to influence the style of generated text in a nuanced and parameterisable way, distinguishing it from prompt engineering. The presented research constitutes a significant step towards developing more adaptive and effective AI-empowered interactive systems. 
\end{abstract}

\section{Introduction}\label{S:Introduction}
Large language models (LLMs) pre-trained on vast corpora have marked a significant milestone in natural language processing, presenting remarkable language understanding and generation capabilities. Models like GPT-2~\citep{radford2019language} and more recent variants such as GPT-3~\citep{brown2020language} and GPT-4~\citep{gpt4} have become influential in transforming the landscape of text generation.
LLMs have the potential to encode extensive public knowledge and can respond to a wide array of text prompts in a manner that often closely resembles human communication. OpenAI's ChatGPT, in particular, has garnered substantial attention, propelling discussions about generative AI from the scientific community into the broader public sphere~\citep{brown2020language,gpt4}. In this era of ever-advancing AI, it is becoming increasingly apparent that LLM-based artificial assistants will play a prominent role in both professional and personal contexts~\citep{bender2021dangers,zhao2023chatgpt}. 
Examples of these are conversational information search~\citep{alessio2023decaf, shah2023taking}, human-AI co-creation~\citep{yuan2022wordcraft, chung2022talebrush}, or complex goal-oriented dialogues~\citep{snell2022context}. 

\begin{figure}[t]
    \centering
    \includegraphics[scale=1]{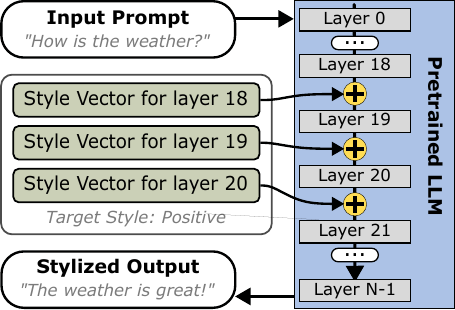}
    \caption{The LLM output is steered by adding style vectors to selected layers (e.g., layers 18-20) during a forward pass. For example, the answer of the LLM to the input prompt ``How is the weather?'' is steered towards a \textbf{positive} style, with a sample answer of ``The weather is great!'', a positive answer.
    }
    \label{fig:steering_process}
\end{figure}

In these complex settings, text generation on a lexical level alone is not sufficient for effective human-AI interaction. Over and above that, a cognitive AI assistant should also be able to adapt to the human user on an affective and emotional level regarding engagement, regulation, decision-making, and discovery~\cite{zhao2022emotion}. There is evidence that LLMs perform well on affective computing tasks, such as sentiment classification and personality prediction, and can have emotional dialogue capabilities to some extent. However, the resulting capabilities do not go far beyond simpler specialized models, presumably due to the LLMs' generality~\citep{zhao2023chatgpt, amin2023}. 
This limitation calls for mechanisms to better control implicit information and the style of an LLM's output.

Prompt engineering has been a promising approach in human-AI collaborative tasks, improving task efficiency and user collaboration~\citep{wu2022ai}. However, it is often highly task-specific and entails manually crafting prompts.

In this paper, we build upon and extend the works of \citet{subramani2022extracting} and \citet{turner2023steering}, which focus on steering the output of LLMs by modifying their internal states. In a series of experiments, using datasets of text samples labeled with sentiments and emotion categories, we show that one can derive a vector representation of a desired style class (e.g., \textit{positive} sentiment) that, when added to the activation of certain layers of an LLM (in this work LLaMa~\citep{touvron2023llama}), its output shows characteristics of this style class (see Fig.~\ref{fig:steering_process}). Our experiments show that the effect of the changed models is more salient when prompted with subjective input (e.g.,``How do you define art?'') rather than with factual input that allows little degrees of freedom (e.g., ``What is the world’s longest river?''). 
Our research aims to bridge the gap between the LLM's capabilities and the nuanced requirements of human-AI interactions, thus extending this novel dimension to the realm of controlling LLM outputs.

An open-source implementation of the algorithms used in this paper is available\footnote{Find all resources at \url{https://github.com/DLR-SC/style-vectors-for-steering-llms}}.

\section{Background and Related Work}\label{S:RelatedWork}
The introduction of transformer architectures in neural networks~\citep{vaswani2017attention} has led to a massive leap in the development of contextualized language models, such as GPT~\citep{brown2020language}. These novel large language models (LLMs) capture relations in the natural data and implicitly encode an unlimited number of more abstract concepts, such as sentiment or style. This quality has been exploited in several recent investigations and can be both a risk~\citep{wagner2022gender} and a chance~\citep{schramowski2022large}. 

Many approaches have been developed with the aim of controlling or affecting the output of LLMs, also referred to as \textit{steering} LLMs \citep{brown2020language,zhang2022survey, jin-etal-2022-deep}. 

Traditionally, methods for producing text in a specific style fall under the domain of \textit{stylized response generation} \citep{sun-etal-2022-stylized, yang-etal-2020-styledgpt, gao-etal-2019-structuring, jin2020hooks}. Nonetheless, as common approaches of this class necessitate training and fine-tuning whole models, these methods are not applicable to state-of-the-art LLMs, given the immense parameter count and training costs of LLMs~\citep{hu2021lora}.

Another line of research worth mentioning that aims to employ alternative approaches to the traditional fine-tuning approach is the parameter-efficient transfer learning approach~\cite{houlsby2019parameter} using adapter modules, which seek to minimize trainable parameters. In contrast, in our work, we focus on a different efficiency aspect, not only on the minimal computational resources but also on the minimal data resources used. 

A related but conceptually different approach to affect the output of LLMs is \textit{text style transfer} (TST) \citep{jin-etal-2022-deep, reif-etal-2022-recipe}. TST aims to transfer the style of a given text into a desired, different style. In contrast, steering LLMs deals with the task of generating a response in a desired style. %
We refer to \citet{jin-etal-2022-deep} for a detailed overview of TST. 

\textit{Prompt engineering} \citep{keskar2019ctrl,radford2019language,shin2020autoprompt,brown2020language,lester2021power,li2021prefix,wei2022chain,wu2022ai} focuses on controlling and directing the output of a language model by designing input prompts or instructions. By tailoring the natural language prompts, the model's output can be steered towards producing responses in the desired style. %

Some recent approaches move in a new direction by modifying the layer activations of an LLM during the forward pass \citep{subramani2022extracting, turner2023steering, hernandez2023measuring}. These approaches can be grouped under the term of \textit{activation engineering}. \citet{subramani2022extracting} presented so-called steering vectors that, when added to the activations at certain layers of an LLM, steer the model to generate a desired target sentence $x$ from an empty input. The rationale behind this is that the information needed to produce the target sentence is already encoded in the underlying neural network. Thus, the approach works without re-training or fine-tuning the model itself. 

Starting with an empty prompt, i.e., beginning of sentence token \textit{<bos>}, the vector $\mathbf{z}_{steer} \in \mathbb{R}^d$ is added to the activations of a defined layer of the model, where $d$ is the dimension of the layer to generate the next of the $T$ tokens of $x$. 
The objective is to find a steering vector $\mathbf{\hat{z}}_{steer}$ that maximizes the log probability: 
\begin{equation}
    \mathbf{\hat{z}}_{steer} = \underset{\mathbf{z}_{steer}}{argmax} \sum_{t=1}^T log ~ p(x_t|x_{<t}, z_{steer}) 
\label{eq:zsteer}
\end{equation}
It was demonstrated on a subset of sentences of the Yelp Sentiment dataset~\citep{shen2017style} that steering vectors can be used for shifting the style of a sentence $x$ towards a dedicated target style using the vector arithmetic: 
\begin{equation}
    \mathbf{\hat{z}}_{target} = \mathbf{z}_{source} + \lambda~\mathbf{z_{\Delta}}
    \label{eq:sentiment_shift}
\end{equation}
$\mathbf{z}_{source}$ is the steering vector that produces sentence $x_{source}$. $\mathbf{z_{\Delta}} = \mathbf{\bar{z}}_{target} - \mathbf{\bar{z}}_{source}$ is the difference between the average of all steering vectors learned for sentences from the target and source domain. The steering vector $\mathbf{\hat{z}}_{target}$ can then be used to steer the model to generate a sentence $x'$ that is similar to $x$ but in the target style.

Moreover, layer activations have demonstrated utility in steering LLMs. \citet{turner2023steering} exemplify that steering vectors, derived from contrasting activations for semantically opposed inputs like ``love'' and ``hate'' can guide LLM outputs during sentence completion. The difference in activations from such contrasting prompts at layer $i$ can straightforwardly be added to another input's activations to steer outputs.

In this work, we add to this line of research a method that efficiently steers LLM outputs towards desired styles with notable control and transparency. In contrast to the aforementioned steering vector and TST techniques, it requires no additional optimization or prior knowledge about original styles.
Unlike prompt engineering, our approach offers quantifiable adjustments in style, providing nuanced differences in responses without relying on vague intensity indicators in prompts, such as ``extremely negative'' versus ``negative.''


\section{Methodology}\label{S:Method}
\begin{figure}[t]
\begin{center}
\centerline{\includegraphics[width=\columnwidth]{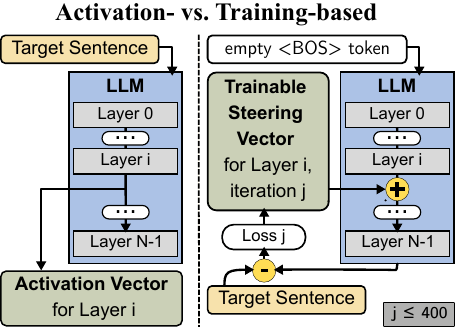}}
\caption{Extraction of an activation vector (left): The LLMs' values at layer $i$ for a prompt in the target style are saved for later computation of style vectors. Trained steering vectors (right): The values of the vectors are optimized over $j=400$ epochs such that the model produces a specified sentence in the target style from a simple beginning of a sentence (BOS) token.
}
\label{fig:SV}
\end{center}
\vskip -0.2in
\end{figure}

We aim to modify the LLM activations for an input $x$ to generate an output that is steered towards a specific style category $s \in S$. As shown in Eq.~\ref{eq:steering}, this is achieved by finding style vectors $\mathbf{v}^{(i)}_{s}$ associated to $s$ such that when added to the activations $\mathbf{a}^{(i)}(x)$ at layer $i$  the output becomes steered towards $s$.
\begin{equation}
\mathbf{\hat{a}}^{(i)}(x) = \mathbf{a}^{(i)}(x) + \lambda\mathbf{v}^{(i)}_{s}
    \label{eq:steering}
\end{equation}

Style categories can be, for example, \emph{positive} and \emph{negative} for sentiment styles or different emotion classes such as \textit{joy} and \textit{anger}. The weighting parameter $\lambda$ (Eq.~\ref{eq:steering}) determines the influence strength of the style vector on the model's output and, thus, allows for more nuanced and controllable model steering compared to prompt engineering. 

In this study, we compare two main approaches to calculate style vectors, namely \textit{Training-based Style Vectors}~(Sec.~\ref{SS:Meth:trained_sv}) and \textit{Activation-based Style Vectors}~(Sec.~\ref{SS:Meth:activation_sv}). Training-based style vectors are found from the generative steering vectors \citep{subramani2022extracting}. In contrast to this generative approach, activation-based style vectors are found by aggregating layer activations for input sentences from the target style \citep{turner2023steering}. The basic assumption behind this is that LLMs internally adapt to the style of the input prompt when producing output, and thus, style vectors can be derived from its hidden states. These two methods are contrasted in Fig.~\ref{fig:SV} and introduced in more detail in this section.

\subsection{Training-based Style Vectors} \label{SS:Meth:trained_sv}

In the approach of \citet{subramani2022extracting} (see Sec.~\ref{S:RelatedWork}), an individual steering vector is learned for each target sentence. Thus, shifting the $source$ style of an unsteered model output $x$ towards a modified output $x'$ (generated by steering vector $\mathbf{\hat{z}}_{x'}$) in the desired $target$ style requires to compute a steering vector $\mathbf{z}_x$ that leads the unconditioned model to produce $x$ (Eq.~\ref{eq:sentiment_shift}). 
This, however, leads to high computational costs and is impractical for online adaptation of an LLM prompted with arbitrary inputs. Furthermore, this vector arithmetic only works for style shifts when the source style is known.
Many styles, such as emotions, have multiple categories. For $n$ style classes, one would need to build $n \times (n-1)$ contrasting vectors $\mathbf{\bar{z}}_{target} - \mathbf{\bar{z}}_{source}$. Consequently, style-shifting is limited and does not generalize to more complex style concepts. 

\paragraph{Our adaptation}
In contrast to the approach of \citet{subramani2022extracting}, we do not shift output styles on sentence level from \textit{source} to \textit{target}. Instead, the steering vectors $\mathbf{z}_x$ learned to steer the model to generate a sample $x$ from style category $s$ are mean-aggregated into a vector $\mathbf{\bar{z}}_{s}^{(i)}$ and all other steering vectors are mean-aggregated into a vector $\mathbf{\bar{z}}_{S\backslash s}^{(i)}$. 
Style vectors $v_s^{(i)}$ for different layers $i$ can then be calculated as in Eq.~\ref{eq:trained_sv}.

\begin{equation}
    \mathbf{v}^{(i)}_s = \mathbf{\bar{z}}_{s}^{(i)} - \mathbf{\bar{z}}_{S \backslash s}^{(i)}
\label{eq:trained_sv}
\end{equation}

Using the average steering vector $\mathbf{\bar{z}}_{S \backslash s}$ as an offset has the advantage that no knowledge about the source style is required to steer the produced output towards a target style.

The training of an individual steering vector $\mathbf{z}_x$ is presented in the right part of Fig.~\ref{fig:SV}. The process begins with the frozen model receiving an empty input token and a steering vector initialized randomly to initiate sentence generation. The resulting output is then evaluated against the target sentence to calculate a cross-entropy loss, which is back-propagated to learn the steering vector.
The training for an output $x$ terminates when a steering vector $\mathbf{z}_x$ that produces the target sentence $x$ is found or after a maximum number of $j=400$ epochs. We use the Adam optimizer~\citep{kingma2014adam} with a learning rate of $0.01$.

\subsection{Activation-based Style Vectors} \label{SS:Meth:activation_sv}
An alternative to relying on trained steering vectors is to work solely in the space of layer activations when the model is prompted with samples from a style category $s$ as suggested by \citet{turner2023steering} (see left-hand side of Fig.~\ref{fig:SV}). 
However, the effect of this approach on the model output has only been shown to be able to steer the output of an LLM for pairs of natural-language prompts by contrasting the activations of those (e.g., ``love'' and ``hate''). 
In this work, we take up this idea and extend it to calculating general style vectors associated with style categories instead of single pairs. 

\paragraph{Our adaptation}
The vector of activations of layer $i$ of an LLM for input $x$ is given as $\mathbf{a}^{(i)}(x)$. The mean-aggregated activations of layer $i$ for all sentences from style category $s \in S$ is denoted as $\mathbf{\bar{a}}^{(i)}_{s}$. 
Analogous to the procedure of Sec.~\ref{SS:Meth:trained_sv}, activation-based style vectors for style category $s$ are calculated as:
\begin{equation}
    \mathbf{v}^{(i)}_{s} =  \mathbf{\bar{a}}^{(i)}_{s} - \mathbf{\bar{a}}^{(i)}_{S \backslash s} 
\label{eq:activations}
\end{equation}
The advantage of this approach is that style vectors are solely based on aggregated activations of chosen layers that are recorded during the forward pass of a sentence of class $s$, and no costly training of steering vectors is required. 


\section{Experiments}\label{S:Experiments}
We compare both introduced approaches, i.e., \textit{training-based style vectors} (Sec.~\ref{SS:Meth:trained_sv}) and \textit{activation-based style vectors} (Sec.~\ref{SS:Meth:activation_sv}) in terms of how well they encode information about style (Sec.~\ref{SS:Results_Probing}) and the ability to steer the model's output (Sec.~\ref{SS:Results_Texts}). 

\subsection{Datasets for Style Definitions}\label{SS:Meth:DataSets}
Experiments are performed along different style categories: sentiment, emotion, and writing style (modern vs. Shakespearean). Each style category is defined through datasets with labeled samples. All datasets used contain English text only. 
For the training-based style vectors, we filter out samples containing more than 50 characters from each dataset to keep the time for computing steering vectors feasible. For details, see Sec.~\ref{SS:Exp:Models}. This limitation does not apply to the activation-based style vectors. 

For our experiments, we use the following popular datasets:

\paragraph{Yelp Review Dataset}
The dataset~\citep{shen2017style} contains unpaired data about restaurant reviews on the Yelp platform labeled as \textit{positive} or \textit{negative}. 
After dropping duplicates, the dataset contains 542k samples. 

\paragraph{GoEmotions}
As a multi-class style dataset, the GoEmotions dataset~\citep{demszky2020goemotions} comprises $58k$ manually curated user comments from the internet platform Reddit\footnote{Reddit forum: \url{https://www.reddit.com/}} labeled with 27 emotional categories. We use $5k$ samples that can be unambiguously mapped to the established six basic emotion categories \citep{Ekman1992}: \emph{sadness}, \emph{joy}, \emph{fear}, \emph{anger}, \emph{surprise}, and \emph{disgust}. 

\paragraph{Shakespeare} 
The Shakespeare dataset \citep{jhamtani-etal-2017-shakespearizing} contains paired short text samples of Shakespearean texts and their modern translations. We use the training set containing 18,395 sentences for each style: modern and Shakespearean.

\subsection{Experimental Setup}\label{SS:Exp:Models}
The aim is to investigate the ability to influence the style of an LLM in a setting where an answer to a question or instruction prompt is expected. Our experiments utilize the open-source Alpaca-7B~\citep{alpaca} ChatGPT alternative, which is based on Meta's LLaMA-7B~\citep{touvron2023llama} architecture. Choosing this model resulted in $d=4096$-dimensional style vectors for each of its 33 layers. We used a single NVIDIA A100-SXM4-80GB for our experiments.

For the evaluation of the training-based style vectors, we only incorporate steering vectors that reproduce the target sentence with $loss < 5$, as vectors with higher $loss$ tend to yield grammatically incorrect output sentences. This resulted in $470$ vectors per layer for the Yelp review dataset, $89$ for GoEmotions, and $491$ for the Shakespeare dataset. In a pre-study on a smaller subset of the data, we found that the steering vectors for the layers $i \in \{18,19,20\}$ are most effective, which is supported by the findings of our probing study (Sec.~\ref{SS:Results_Probing}). We only train steering vectors for these layers to keep the computational effort feasible. Nevertheless, we had to run the experiment on the Yelp and Shakespeare datasets for 150 hours each and for GoEmotions for around 100 hours. In comparison, the extraction of the activations only took at most 8 hours per dataset and resulted in recorded activation vectors for all dataset samples. 

\subsection{Probing Study}\label{SS:Results_Probing}

\begin{figure*}[th]
     \centering
     \begin{subfigure}[t]{0.329\textwidth}
        \captionsetup{justification=centering}
         \centering
         \includegraphics[width=\linewidth]{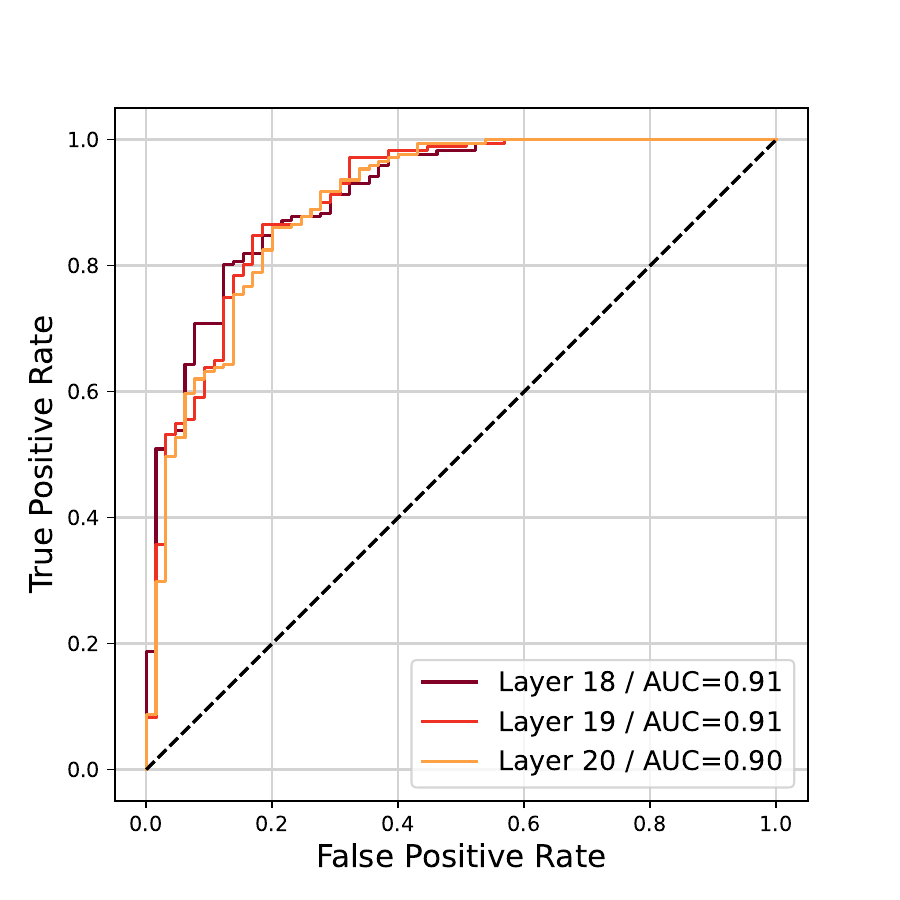}
         \caption{Trained steering vectors}
         \label{fig:roc_yelp_steering}
     \end{subfigure}
     \hfill
     \begin{subfigure}[t]{0.329\textwidth}
        \captionsetup{justification=centering}
         \centering
         \includegraphics[width=\linewidth]{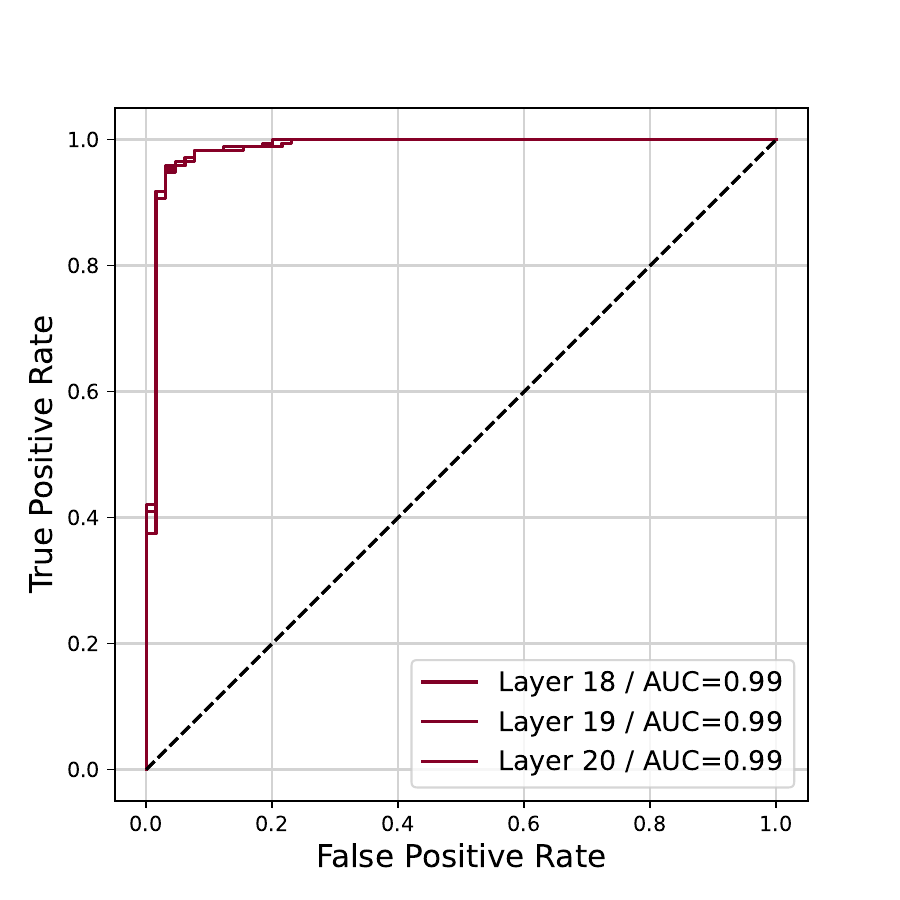}
         \caption{Corresponding activation vectors}
         \label{fig:roc_yelp_acti_fair}
     \end{subfigure}
     \hfill
     \begin{subfigure}[t]{0.329\textwidth}
        \captionsetup{justification=centering}
         \centering
         \includegraphics[width=\linewidth]{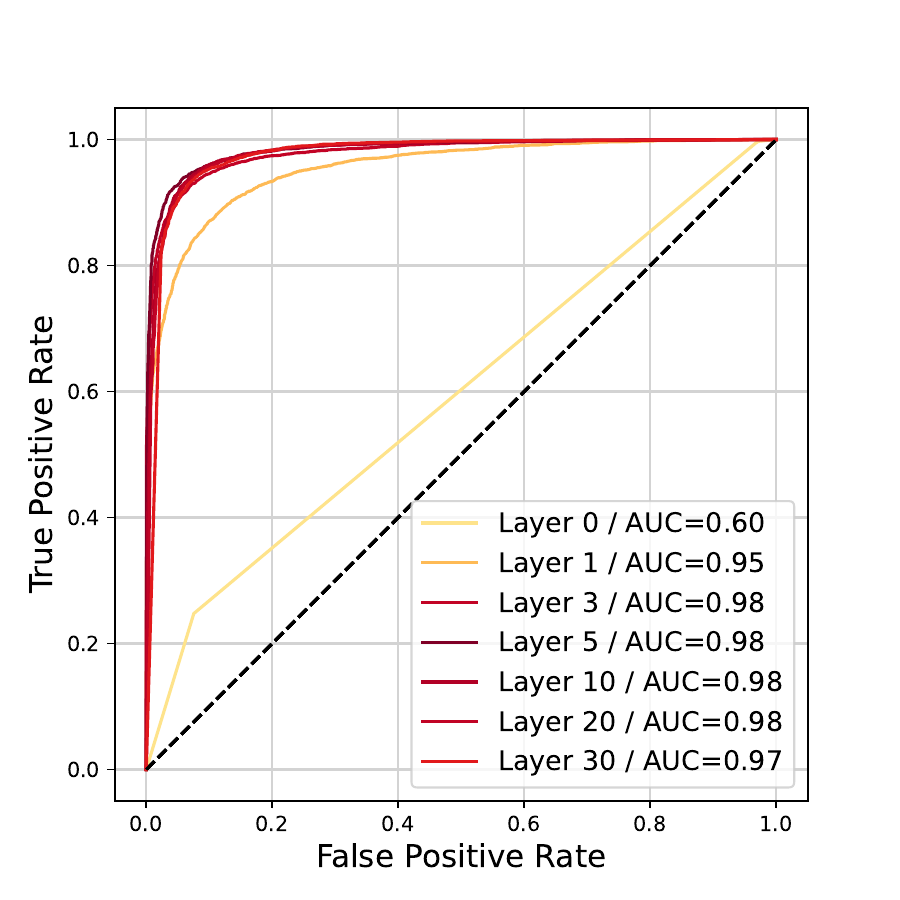}
         \caption{Activation vectors of 10k sentences}
         \label{fig:roc_yelp_acti_all}
     \end{subfigure}
        \caption{Classification results on the Yelp review dataset: Using (a) only the 470 trained steering vectors, (b) the corresponding activation vectors, and (c) selected layers of activation vectors of 10k sentences. The activation vectors show superior performance in their ability to predict the sentiment of an input sentence.}
        \label{fig:roc_curves_yelp}
\end{figure*}

The receiver operating characteristic (ROC) curves for two class predictions (positive and negative sentiment) in the Yelp review dataset are presented in Fig.~\ref{fig:roc_curves_yelp}. It can be seen that, in general, activations from layer three onwards lead to remarkably high classification accuracy (AUC $\geq 0.97$, see Fig.~\ref{fig:roc_yelp_acti_all}) and are almost perfect for layers $i \in \{18,19,20\}$. 
As expected, activations encode style more explicitly than trained steering vectors, which still achieve considerable accuracy. The results are similar for the other two datasets, discussed in Sec.~\ref{S:Further_Probing}.

We can, therefore, determine that the layers $i \in \{18,19,20\}$ are candidates for effective steering, and we only use style vectors $\mathbf{v^{(i)}}_{s}$ computed from these layers for the generation of prompts in the next section.

\subsection{Evaluation of Generated Texts}
\label{SS:Results_Texts}

\begin{figure*}[t]
     \centering
     \begin{subfigure}[t]{0.329\textwidth}
        \captionsetup{justification=centering}
         \centering
         \includegraphics[width=\linewidth]{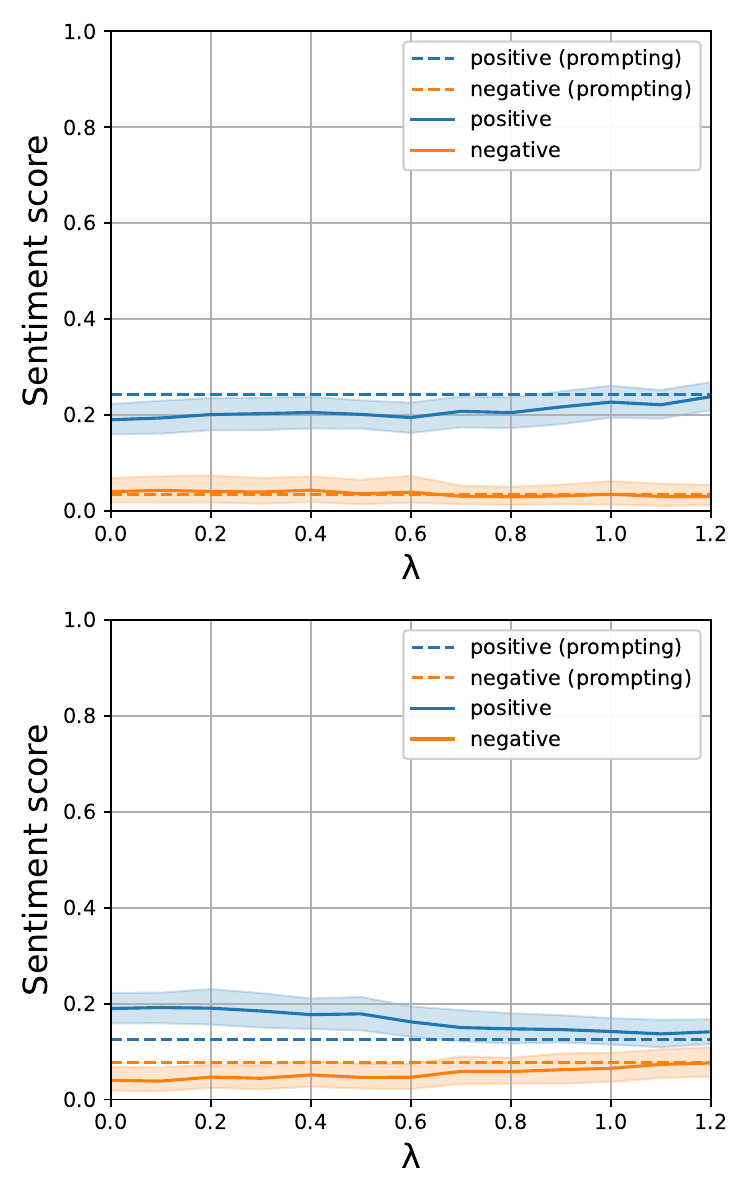}
         \caption{Style vectors from trained steering vectors}
         \label{fig:yelp_contrastive_subjective_source_original_trained_vector_based_lda1}
     \end{subfigure}
     \hfill
     \begin{subfigure}[t]{0.329\textwidth}
        \captionsetup{justification=centering}
         \centering
         \includegraphics[width=\linewidth]{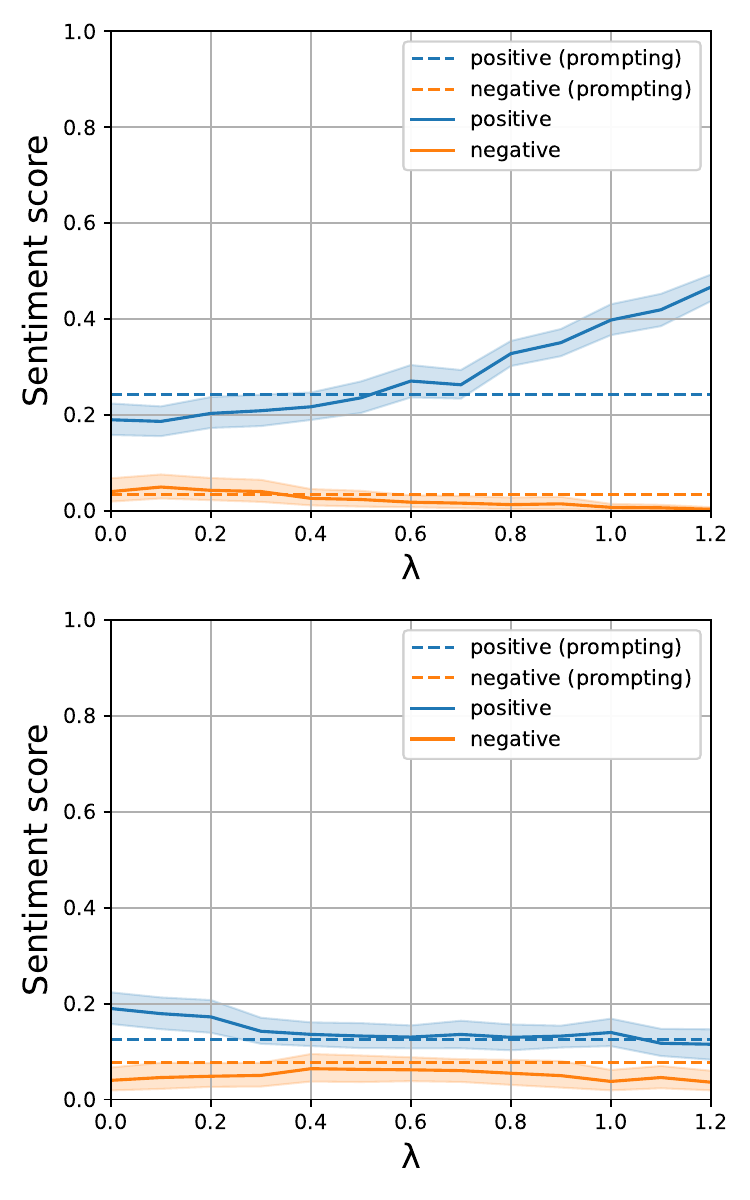}
         \caption{Style vectors from the corresponding activation vectors}
         \label{fig:yelp_contrastive_subjective_source_original_activation_based_lda1}
     \end{subfigure}
     \hfill
     \begin{subfigure}[t]{0.329\textwidth}
        \captionsetup{justification=centering}
         \centering
         \includegraphics[width=\linewidth]{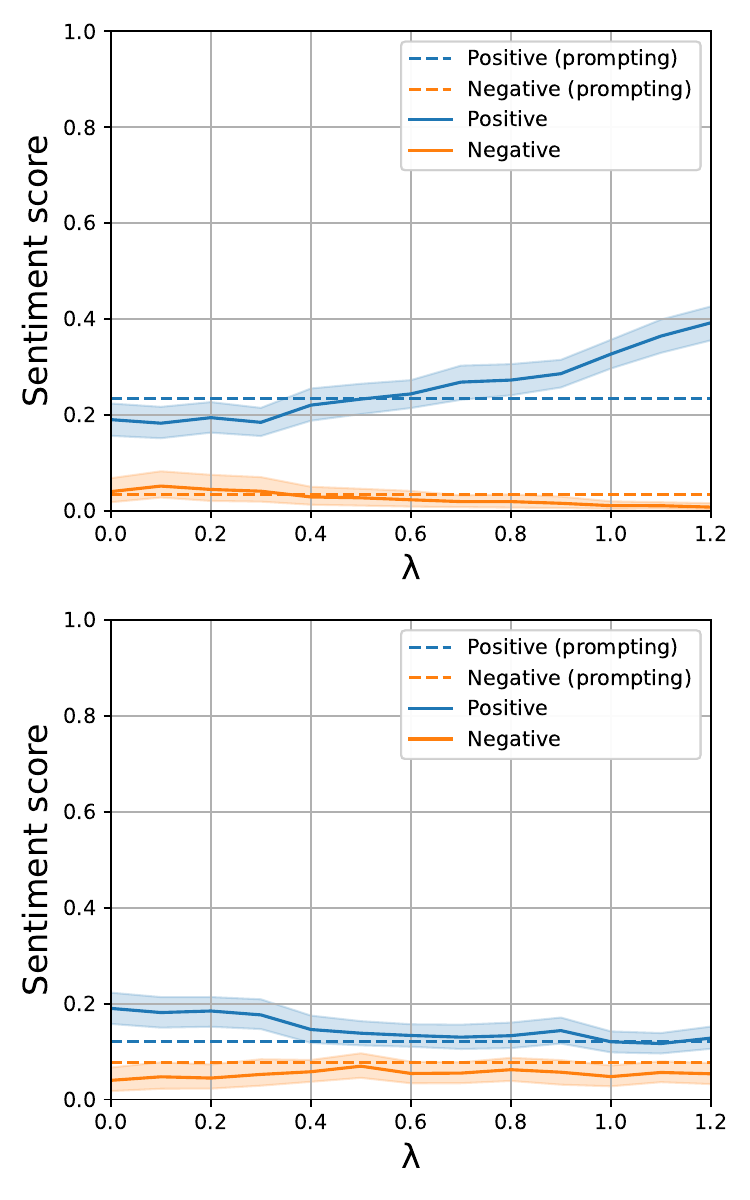}
         \caption{Style vectors from all activation vectors}
         \label{fig:yelp_contrastive_subjective_source_original_activation_based_multi_new_questions_lda1}
     \end{subfigure}
        \caption{Steering of the Yelp Review samples towards positive (upper plots) and negative (lower plots) sentiment. }
        \label{fig:yelp}
\end{figure*}

As shown in Sec.~\ref{SS:Results_Probing}, both trained steering and activation vectors capture relevant style information. However, this does not show that style vectors $\mathbf{v^{(i)}}_{s}$ that are computed from them can be used to actually steer the style of the model's output. For this reason, we assembled a list of 99 exemplary prompts as input for the Alpaca-7B model. Since the style of an LLM's output cannot be considered independently of the type of input prompt, we created two different sets of prompts: The factual list comprises 50 prompts that ask about a hard fact with a clear, correct answer, such as ''\textit{Who painted the Mona Lisa?}``. The subjective list includes 49 different prompts, allowing more individual responses to express sentiments and emotions. They either inquire about a personal opinion, e.g., ''\textit{What do German bread rolls taste like?}``, or general information and allow for a variety of responses, for instance, ''\textit{Describe a piece of artwork}.`` Steering towards a sentiment or emotion category is expected to affect the LLM's outcome significantly more for such subjective prompts than for factual prompts. The full list of prompts is given in Sec.~\ref{S:Evaluation_prompts}.

As described in Section~\ref{S:Method}, the parameter $\lambda$ of Eq.~\ref{eq:steering} influences how strongly the model is steered towards the target style. We found that if this parameter is chosen too large, the model sometimes produces nonsense texts, as shown in Ex. E2 in Sec.~\ref{SSS:Steer_Examples} and in Appendix in Sec.~\ref{S:effect_of_lambda}. This effect seems to be dependent on the input prompt and style domain.

\subsubsection{Classification-based Evaluation}
\label{SSS:classification_based_eval}
\begin{figure*}[t]
     \centering
     \begin{subfigure}[t]{0.329\textwidth}
        \captionsetup{justification=centering}
         \centering
         \includegraphics[width=\linewidth]{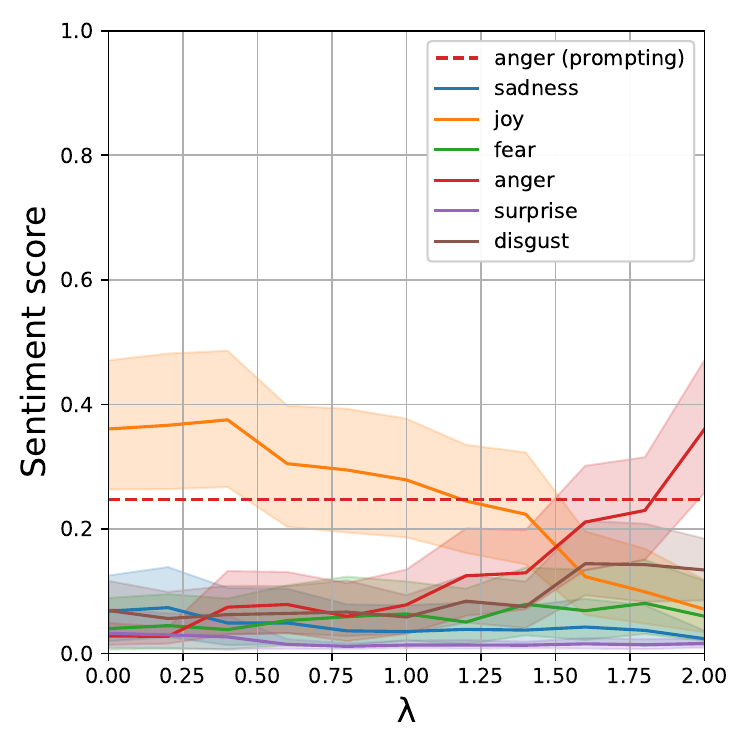}
         \caption{Steering to anger, \\subjective prompts}
         \label{fig:contrastive_steering_anger_subjective_multi}
     \end{subfigure}
     \hfill
     \begin{subfigure}[t]{0.329\textwidth}
        \captionsetup{justification=centering}
         \centering
         \includegraphics[width=\linewidth]{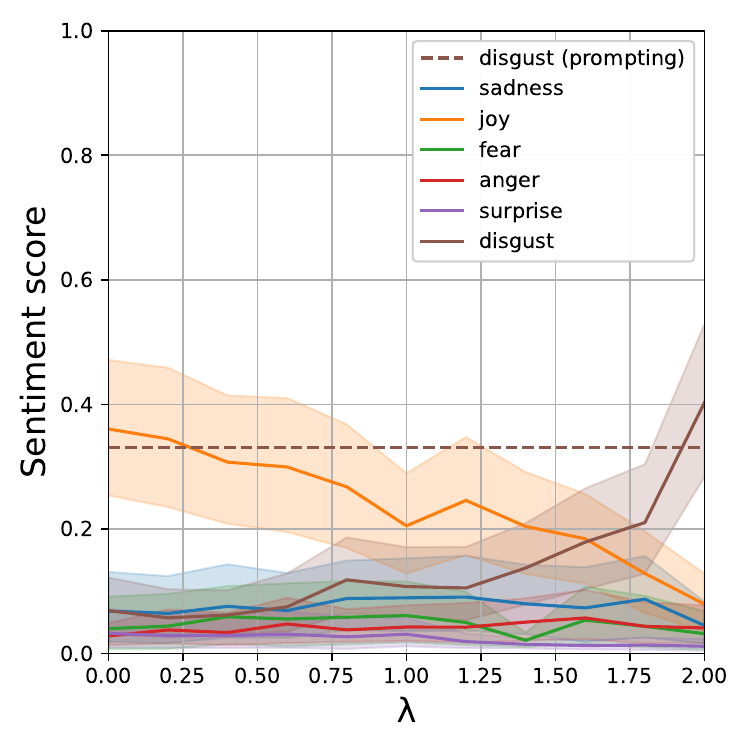}
         \caption{Steering to disgust, \\subjective prompts}
         \label{fig:contrastive_steering_disgust_subjective_multi}
     \end{subfigure}
     \hfill
     \begin{subfigure}[t]{0.329\textwidth}
        \captionsetup{justification=centering}
         \centering
         \includegraphics[width=\linewidth]{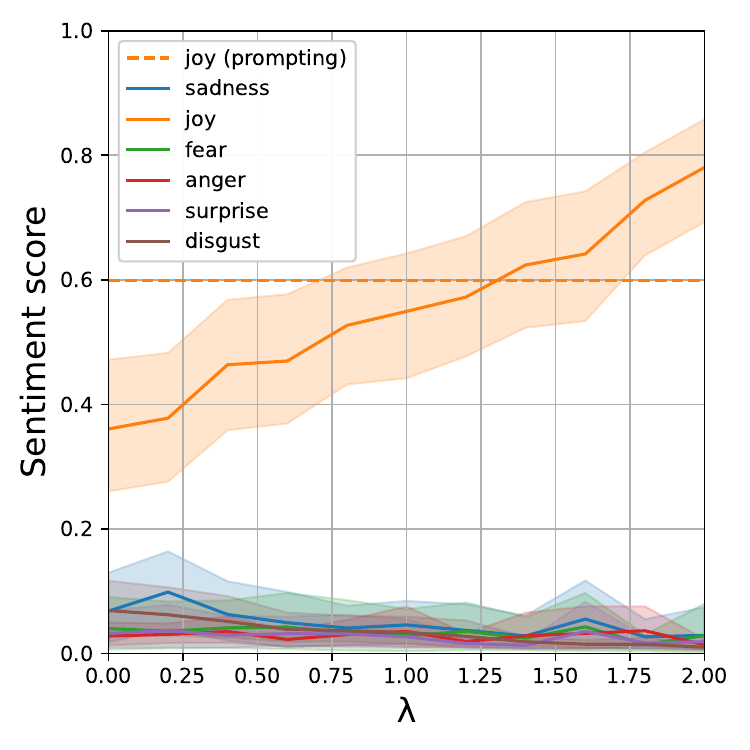}
         \caption{Steering to joy, \\subjective prompts}
         \label{fig:contrastive_steering_joy_subjective_multi}
     \end{subfigure}

          \begin{subfigure}[t]{0.329\textwidth}
        \captionsetup{justification=centering}
         \centering
         \includegraphics[width=\linewidth]{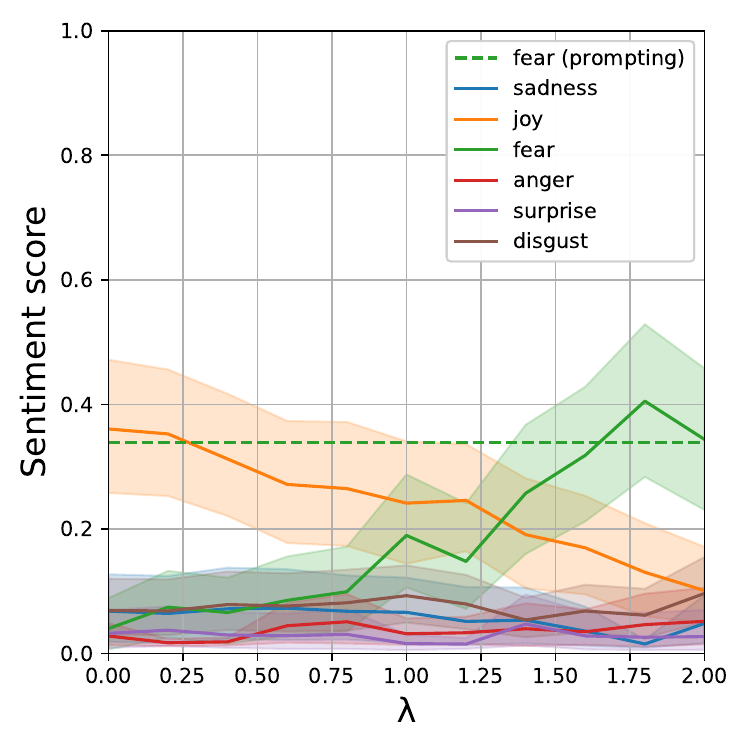}
         \caption{Steering to fear, \\subjective prompts}
         \label{fig:contrastive_steering_fear_subjective_multi}
     \end{subfigure}
     \hfill
     \begin{subfigure}[t]{0.329\textwidth}
        \captionsetup{justification=centering}
         \centering
         \includegraphics[width=\linewidth]{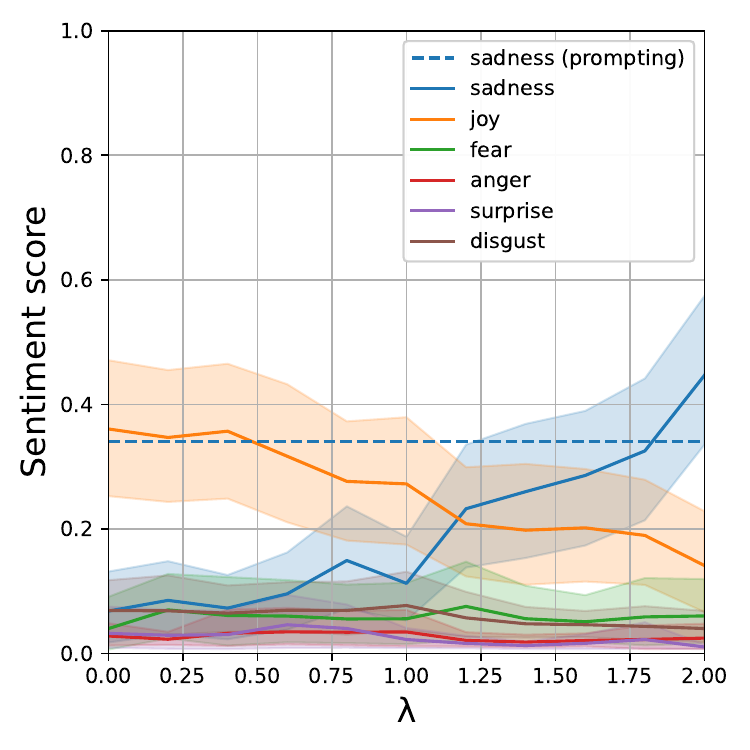}
         \caption{Steering to sadness, \\subjective prompts}
         \label{fig:contrastive_steering_sadness_subjective_multi}
     \end{subfigure}
     \hfill
     \begin{subfigure}[t]{0.329\textwidth}
        \captionsetup{justification=centering}
         \centering
         \includegraphics[width=\linewidth]{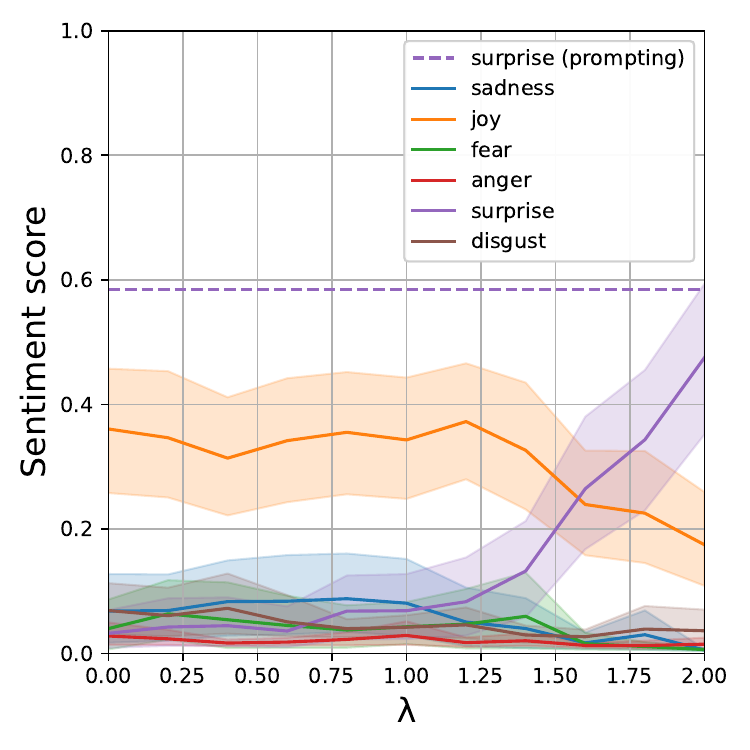}
         \caption{Steering to surprise, \\subjective prompts}
         \label{fig:contrastive_steering_surprise_subjective_multi}
     \end{subfigure}
        \caption{Activation-based style vectors: Evaluation of generated texts for \textit{subjective} prompts using GoEmotions' style vectors. All activation vectors were used.}
        \label{fig:goemo_subjective}
\end{figure*}

We use standard classification models to evaluate the steered output of training and activation-based style vectors. 
The dashed lines in all steering plots, e.g., in Fig.~\ref{fig:yelp} and Fig.~\ref{fig:goemo_subjective}, indicate the mean classification score achieved for a prompting baseline. In these instances, no steering vector was applied to the model. Instead, we appended ``Write the answer in a [\dots] manner.'' to the input prompt, where the dots are replaced with the respective target steering style, e.g., \textit{positive}, or \textit{angry}. Thus, the model is informed in a neutral way to direct the output as required. 

For the Yelp dataset-based style vectors, the positivity and negativity values of produced outputs were inferred by the VADER sentiment analyzer~\citep{Hutto_Gilbert_2014} as a state-of-the-art model. 
Fig.~\ref{fig:yelp} shows the average sentiment classification scores on the model's steered outputs for different values of  $\lambda$ and the $49$ subjective input prompts. It appears that steering into the positive direction works better in general, while the steering effect is stronger for activation-based style vectors. As one could expect, for the 50 factual prompts, there are no notable differences since the factual answers are mostly neutral. Thus, corresponding plots are omitted. The prompt baseline, on average, demonstrates only a minimal effect compared to the model's default output.

In order to evaluate the ability of the style vectors derived from the GoEmotions dataset to steer the model 's output towards multiple emotional categories, a multi-class model based on RoBERTa~\citep{hartmann2022emotionenglish} was used. Fig.~\ref{fig:goemo_subjective} depicts the shift in the classified emotion class score when the style space is built from activation-based style vectors on the GoEmotions dataset. 
The notable increase in the respective emotion class score indicates a strong model performance when steering towards a specific emotion. As expected, styling the output on factual prompts shows almost no changes in the emotion class scores, similar to the results for the Yelp style vectors described above. 
Prompt baselines created with the GoEmo labels exhibit a substantially stronger impact on the generated outcomes compared to the prompt baseline in the case of the Yelp dataset.

\subsubsection{Steering Output Examples}
\label{SSS:Steer_Examples}
In the following, selected examples of model outputs modified by style vectors are presented 
to provide an impression of the obtained results under different settings. In each case, \textbf{P} is the prompt, followed by different model outputs. 

\paragraph{E1 - Opposite sentiment directions.} Examples of how the output changes when shifted in opposite directions using activation-based style vectors derived from the Yelp dataset at $\lambda = 1$. The target style is \textit{positive} for \textbf{O1} and \textit{negative} for \textbf{O2}.

\begin{itemize}\itemsep0em 
   \item[\bf P:] \textit{Share a personal anecdote about a vacation you had}
   \item[\bf O1:] \textit{My favorite vacation was a trip I took to the beach with my family. We had a great time exploring the beach, going on walks, and tasting all the delicious food. The best part was the relaxing atmosphere and the great memories we made.} 
   \item[\bf O2:] \textit{I had a terrible experience with a vacation I had. I went to a beach in a different country and I was expecting to have a great time. Instead, I was met with a huge amount of racism. People were openly saying that I was a “strike-o-bog” (a term that is used to refer to a black person) and that I was “not allowed” to be there. I was then physically attacked. [...]} 
\end{itemize}

\paragraph{E2 - Too large $\lambda$s.} The parameter $\lambda$ in Eq.~\ref{eq:steering} plays a central role for the outputs. Yet, there is no general threshold for $\lambda$, but it depends on the input prompt and the target style definition. Too large $\lambda$ values sometimes result in nonsense outputs with repetitions. This behavior could be observed especially for training-based style vectors. $R$ was the result of steering towards sadness with $\lambda=1$.
\begin{itemize}
    \item[\bf P:] \textit{Describe a fictional character.}
    \item[\bf R:] \textit{Ms. Grey is an elderly woman with a sadly sadly sadly sadly sadly [...].}
\end{itemize}

\paragraph{E3 - Target Bias.}
In reality, datasets do not only encode desired target styles but also other domain-specific characteristics. Style vectors from Yelp restaurant reviews often seem to carry a bias towards food topics in addition to sentiment. 
\textbf{R} results from activation-based steering towards positive style with $\lambda=1.2$.
\begin{itemize}
    \item[\bf P:] \textit{What did a day in a typical family in the year 1980 look like?}
    \item[\bf R:] \textit{A day in a typical family in the year 1980 was filled with great food, fun and memorable moments. The morning started with a delicious breakfast and great conversations. The lunch was always a great home-made meals and the dinner was always a great variety of flavours.}  
\end{itemize}

\paragraph{E4 - Steering Emotion.}
Also, the underlying emotion in the LLM's response can steer, as demonstrated in this example. \textbf{J} and \textbf{A} were steered activation-based towards joy and anger, respectively, with $\lambda=0.8$.
\begin{itemize}
    \item[\bf P:] \textit{How would you respond to a rude customer in a customer service role?}
    \item[\bf J:] \textit{I apologize for the inconvenience. I understand your frustration and thank you for bringing this to my attention. I'd like to help you resolve the issue as quickly as possible. [...]}  
    \item[\bf A:] \textit{I understand you are frustrated, but I cannot allow you to speak to me in this manner. I am here to help you, but you must treat me with respect. [...]} 
\end{itemize}

\paragraph{E5 - Steering style beyond sentiment and emotions.} These outputs were shifted with activation-based style vectors calculated upon the Shakespeare dataset. While \textbf{M} was steered towards \textit{modern} with $\lambda=0.8$, \textbf{S} was steered towards \textit{Shakespearean} with $\lambda=1.6$. 

\textbf{S} is formulated in a more flowery and antiquated language. Presumably, the maximal $\lambda$ for shifting towards \textit{modern} is smaller because this style is per se more similar to the LLM's style and, therefore, also lies closer in the latent vector space.
\begin{itemize}
    \item[\bf P:] \textit{How do you define happiness?}
    \item[\bf M:] \textit{Happiness is a state of contentment, joy, and satisfaction in life. It is the feeling of being satisfied with who you are and having a sense of purpose and fulfillment in life.}
    \item[\bf S:] \textit{Happiness is a state of contentment and joy, wherein the soul is freed from the bondage of sorrow, mischievous fancies, and unworthy thoughts, and wherein the body is freed from the bondage of pain, and wherein the soul duly commends itself to the Lord, and is in some measure made partaker of the blessedness which is past, which is present, or which to come.}  
\end{itemize}

\section{Discussion and Conclusion}\label{S:Discussion}
This work investigated vector representations associated with sentiments, emotion categories, and general writing styles that can influence the output style of LLMs. 
In a generative approach, style vectors were derived from steering vectors found in a training procedure and steered the model to produce samples in a desired style from scratch. In contrast, activation-based style vectors are derived from the activations of input prompts, which relies on the assumption that LLMs internally adapt the input style during the forward pass. Steering vector training is much more expensive than simply recording the hidden layer activation during a single forward pass. Therefore, the activation-based style vectors are the preferred approach for steering style in large language models, both in terms of performance and resource efficiency.

We also found that, for factual prompts, the output can only marginally be influenced. It can be considered positive that one cannot easily dissuade the model from answering in a neutral tone to a factual prompt while still being adaptable if the input permits, especially in conversational settings.

Style vectors enable a continuous and adjustable modulation of the outputs of large language models. Unlike prompt engineering, which offers more step-wise control over style intensities (like ``Write the answer in a positive way'' versus ``Write the answer in a \textit{very} positive way''), style vectors provide smoother transitions. This activation-based control is achievable because the vectors in activation engineering are constructed from known datasets. In contrast, traditional prompting may trigger activations that are unknown and inaccessible to the user, limiting the ability to fine-tune the output.
Furthermore, activation-based steering has the potential to generate new styles, expanding the possibilities beyond the constraints of pre-training knowledge inherent in prompt engineering. While prompt engineering relies on existing knowledge and often involves a trial-and-error approach, activation engineering opens up new avenues for style generation and customization. More complex styles, such as multidimensional composed styles, present unique challenges when approached through activation engineering. However, the advantages it offers, such as enhanced control over the output and the capacity to develop unique styles, significantly outweigh these initial challenges. It is important to note that these methods are not mutually exclusive; they can be combined to leverage each approach's strengths, enhancing our model's overall capability and flexibility.

To the best of our knowledge, this is one of the first studies on steering language models beyond GPT-2 (in our case Alpaca-7B~\citep{alpaca}). Results should, however, be transferable to any other type of LLM with direct access to hidden layer activations. How to determine the exact influence of the weighting parameter $\lambda$ (Eq.~\ref{eq:steering}) is still an open question. $\lambda$ allows for nuanced style steering but, if chosen too large, leads the model to produce nonsense texts. Moreover, this seems to depend on the domain (sentiment, emotion, writing style). We leave this for future research. 

\section*{Limitations}
It was not feasible to derive trained steering vectors for all considered samples since training involves high computational costs and requires a maximal sample length of 50 characters. In contrast, activation-based style vectors could straightforwardly be obtained for every text sample without restrictions. We conducted activation-based experiments on the complete sample set to explore the proposed approach fully. 
However, to avoid a potential bias towards activation-based style vectors and provide a fair comparison, we also conducted our experiments on the subset of samples that could be considered for both settings. 

We evaluated the ability to influence the style of an LLM's output with style vectors using existing sentiment and emotion classifiers. Both classifiers are widely used in practice and have shown state-of-the-art results. However, they are not perfect, and thus, results only show a general tendency. In the future, we plan to conduct studies on individual human perceptions of the text style produced by steered LLMs.

The experiments have a strong focus on sentiment and emotion as style characteristics. Results on the Shakespeare dataset provide evidence that the output of LLMs can also generally be steered towards tone and writing style. This, however, has to be investigated in more depth in the future, especially concerning texts in languages other than English. 

\section*{Ethics Statement}
Our method may generate negative, rude, and hateful sentences about a specific person or a commercial site caused by the data distribution of Yelp and GoEmotions datasets. Therefore, it could be used with malicious intentions, i.e., by targeted harassment or inflation of positive reviews. Since our work involves a pre-trained generative LLM, which was trained on text scraped from the web, it has acquired some biases that were present there. Such biases might be extracted by certain prompts and could even be strengthened by our style steering. Furthermore, it is important to note that steering the style of LLMs may bear the potential to mimic a specific style of speech from persons whose statements were used to train the model. Therefore, the approaches could be abused to create realistic fake statements. 

In the context of image generation, the idea of shifting entities in the latent space during the generation process has already been implemented successfully~\citep{brack2022stable} and can considerably reduce harmful content in generated images~\citep{schramowski2023safe}. Analogously, our approach can also be used to reduce harmful output.  

\section*{Acknowledgements}
The authors gratefully acknowledge the computational and data resources provided through the joint high-performance data analytics (HPDA) project “terrabyte” of the German Aerospace Center (DLR) and the Leibniz Supercomputing Center (LRZ).

\bibliography{references}

\appendix

\section*{Appendix}

\section{Evaluation Prompts}
\label{S:Evaluation_prompts}
In this investigation, we compared the system's performance on \textit{factual} and \textit{subjective} on prompts. Comprehensive lists of these prompts are provided in Sec.~\ref{SS:ASS_factual_prompts} and Sec.~\ref{SS:ASS_subjective_prompts}, respectively.

\subsection{Factual Prompts}\label{SS:ASS_factual_prompts}

There were 50 factual prompts used in this study, which are referred to as \textbf{F01} to \textbf{F50}:
\begin{description}\itemsep0em 
\item{\bf[F01]} How many bones are there in the human body?
\item{\bf[F02]} How many chambers are there in the human heart?
\item{\bf[F03]} How many elements are there in the periodic table?
\item{\bf[F04]} How many planets are there in our solar system?
\item{\bf[F05]} How many players are there in a baseball team?
\item{\bf[F06]} How many players are there in a volleyball team?
\item{\bf[F07]} How many symphonies did Ludwig van Beethoven compose?
\item{\bf[F08]} In which year did World War II end?
\item{\bf[F09]} In which year did the Berlin Wall fall?
\item{\bf[F10]} In which year did the first moon landing occur?
\item{\bf[F11]} What is the boiling point of water in Fahrenheit?
\item{\bf[F12]} What is the capital city of France?
\item{\bf[F13]} What is the chemical formula for methane?
\item{\bf[F14]} What is the chemical formula for table salt?
\item{\bf[F15]} What is the chemical formula for water?
\item{\bf[F16]} What is the chemical symbol for gold?
\item{\bf[F17]} What is the chemical symbol for sodium?
\item{\bf[F18]} What is the deepest point in the Earth's oceans?
\item{\bf[F19]} What is the formula for calculating density?
\item{\bf[F20]} What is the formula for calculating the area of a circle?
\item{\bf[F21]} What is the formula for calculating the area of a triangle?
\item{\bf[F22]} What is the formula for calculating the volume of a cylinder?
\item{\bf[F23]} What is the formula for converting Celsius to Fahrenheit?
\item{\bf[F24]} What is the freezing point of water in Kelvin?
\item{\bf[F25]} What is the largest country in the world by land area?
\item{\bf[F26]} What is the largest internal organ in the human body?
\item{\bf[F27]} What is the largest ocean in the world?
\item{\bf[F28]} What is the largest organ in the human body?
\item{\bf[F29]} What is the speed of light in a vacuum?
\item{\bf[F30]} What is the symbol for the chemical element iron?
\item{\bf[F31]} What is the tallest building in the world?
\item{\bf[F32]} What is the tallest mountain in the world?
\item{\bf[F33]} What is the world's longest river?
\item{\bf[F34]} Which country is famous for the Taj Mahal?
\item{\bf[F35]} Which country is known as the Land of the Rising Sun?
\item{\bf[F36]} Which gas is known as laughing gas?
\item{\bf[F37]} Which gas makes up the majority of Earth's atmosphere?
\item{\bf[F38]} Who developed the theory of evolution by natural selection?
\item{\bf[F39]} Who discovered penicillin?
\item{\bf[F40]} Who discovered the theory of general relativity?
\item{\bf[F41]} Who is considered the father of modern physics?
\item{\bf[F42]} Who is credited with inventing the telephone?
\item{\bf[F43]} Who is the author of the play ``Romeo and Juliet''?
\item{\bf[F44]} Who is the current President of the United States?
\item{\bf[F45]} Who painted ``The Starry Night''?
\item{\bf[F46]} Who painted the ``Last Supper''?
\item{\bf[F47]} Who painted the Mona Lisa?
\item{\bf[F48]} Who wrote the novel ``Pride and Prejudice''?
\item{\bf[F49]} Who wrote the novel ``To Kill a Mockingbird''?
\item{\bf[F50]} Who wrote the play ``Hamlet''?
\end{description}

\subsection{Subjective Prompts}\label{SS:ASS_subjective_prompts}
The 49 applied factual prompts are referred to as \textbf{S01} to \textbf{S49}: 
\begin{description}\itemsep0em 
\item{\bf[S01]} Announce the weather forecast for the upcoming weekend.
\item{\bf[S02]} Ask your hairdresser for an appointment next week to have your hair dyed.
\item{\bf[S03]} Comment on a critical review of a customer of your business.
\item{\bf[S04]} Compare the color blue and green.
\item{\bf[S05]} Compare the cultural value of theaters and cinemas.
\item{\bf[S06]} Compare the qualities of coffee and tea.
\item{\bf[S07]} Compare the relaxation based on vacation and continuous sport.
\item{\bf[S08]} Compare the taste of a strawberry smoothie to that of a vanilla one.
\item{\bf[S09]} Compose a few lines of lyrics talking about society.
\item{\bf[S10]} Describe a fictional character.
\item{\bf[S11]} Describe a meal or dish that holds sentimental value to you and why.
\item{\bf[S12]} Describe a person who has had an impact on your life and why.
\item{\bf[S13]} Describe a piece of artwork.
\item{\bf[S14]} Describe an incident that could lead to an airplane crash in mid-flight.
\item{\bf[S15]} Discuss the impact of social media on interpersonal relationships.
\item{\bf[S16]} How can I learn about Machine Learning most efficiently?
\item{\bf[S17]} How do caterpillars turn into butterflies?
\item{\bf[S18]} How do you approach decision-making when faced with multiple options?
\item{\bf[S19]} How do you define art?
\item{\bf[S20]} How do you define happiness?
\item{\bf[S21]} How do you define sadness?
\item{\bf[S22]} How do you feel about the death penalty?
\item{\bf[S23]} How do you prioritize your tasks and responsibilities in your daily life?
\item{\bf[S24]} How do you stay motivated and focused on long-term goals?
\item{\bf[S25]} How would you handle a disagreement with a close friend?
\item{\bf[S26]} How would you respond to a rude customer in a customer service role?
\item{\bf[S27]} If a roommate consistently borrows your belongings without asking, how would you handle it?
\item{\bf[S28]} Order a vegan dish from the menu of a steak house.
\item{\bf[S29]} Review the pair of headphones that I bought online last week.
\item{\bf[S30]} Share a personal anecdote about a vacation you had.
\item{\bf[S31]} Share a personal experience of overcoming a fear or facing a challenge.
\item{\bf[S32]} Share a personal experience that challenged your beliefs and changed your perspective.
\item{\bf[S33]} Share a quote or mantra that inspires you and explain why it resonates with you.
\item{\bf[S34]} Tell me a joke.
\item{\bf[S35]} Tell me about the concert in America last year.
\item{\bf[S36]} What approach would you take when faced with a challenging work colleague?
\item{\bf[S37]} What approaches do you take to foster creativity and innovation in your work?
\item{\bf[S38]} What are the steps to bake a perfect pizza from scratch?
\item{\bf[S39]} What did a day in a typical family in the year 1980 look like?
\item{\bf[S40]} What do German bread rolls taste like?
\item{\bf[S41]} What does a fulfilling and meaningful life mean to you?
\item{\bf[S42]} What is your opinion on the role of government in society?
\item{\bf[S43]} What is your perspective on the significance of cultural heritage?
\item{\bf[S44]} What is your preferred method of learning new skills or acquiring knowledge?
\item{\bf[S45]} What is your stance on the role of machine learning in education?
\item{\bf[S46]} What strategies do you find most effective for managing stress?
\item{\bf[S47]} Would you like to see a movie about yourself?
\item{\bf[S48]} Write a recipe to make chocolate chip muffins.
\item{\bf[S49]} Write an introduction about yourself for a CV.
\end{description}

The full list of produced answers to the prompts in Sec.~\ref{SS:ASS_factual_prompts} and Sec.~\ref{SS:ASS_subjective_prompts} can be found at [redacted for peer-review].

\section{Effect of the parameter $\lambda$}
\label{S:effect_of_lambda}

The example below shows an example of the effect of the weighting parameter lambda for style vectors (Eq.~\ref{eq:steering}) on steering a response toward negative sentiment. Larger values generally lead to increased negativity. However, if the values are too large, the model's output is not adequate anymore (oversteering).

\begin{itemize}
\item[\bf P:] \textit{Review the pair of headphones that I bought online last week.}
\setlength{\itemindent}{2em}
\item[\bf $\lambda = 0$:] \textit{I recently bought a pair of headphones online and I am very pleased with them. The sound quality is excellent and they are very comfortable to wear. The design is also stylish and the adjustable headband allows for a custom fit. I would highly recommend these headphones to anyone looking for a good pair of audio headphones.}  
\item[\bf $\lambda = 0.6$:] \textit{I recently bought a pair of headphones online and I'm very satisfied with them. The sound quality is great and they are very comfortable to wear. The design is also very stylish and I love the adjustable fit. Highly recommended!} 
\item[\bf $\lambda = 1.2$:] \textit{I love the pair of headphones you got! The sound quality is great and the comfortable design is perfect for all day listening. High recommend!}
\item[\bf $\lambda = 1.9$:] \textit{<s>  great great great great great great ...}
\end{itemize}

\section{Further results from the probing study}
\label{S:Further_Probing}

Analogously to the analysis of the Yelp dataset in Sec.~\ref{SS:Results_Probing}, we performed the same experiment with the Shakespeare and the GoEmotions datasets. 

\begin{figure*}[h]
     \centering
     \begin{subfigure}[t]{0.329\textwidth}
        \captionsetup{justification=centering}
         \centering
         \includegraphics[width=\linewidth]{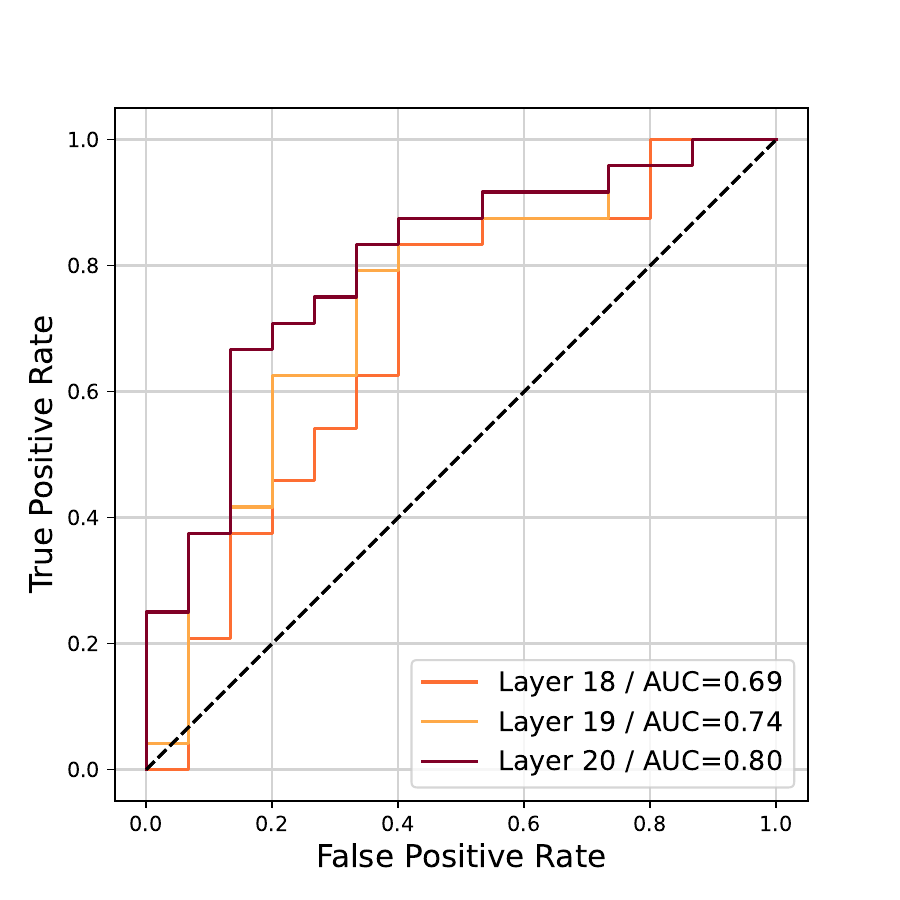}
         \caption{Trained steering vectors}
         \label{fig:roc_shake_steering}
     \end{subfigure}
     \hfill
     \begin{subfigure}[t]{0.329\textwidth}
        \captionsetup{justification=centering}
         \centering
         \includegraphics[width=\linewidth]{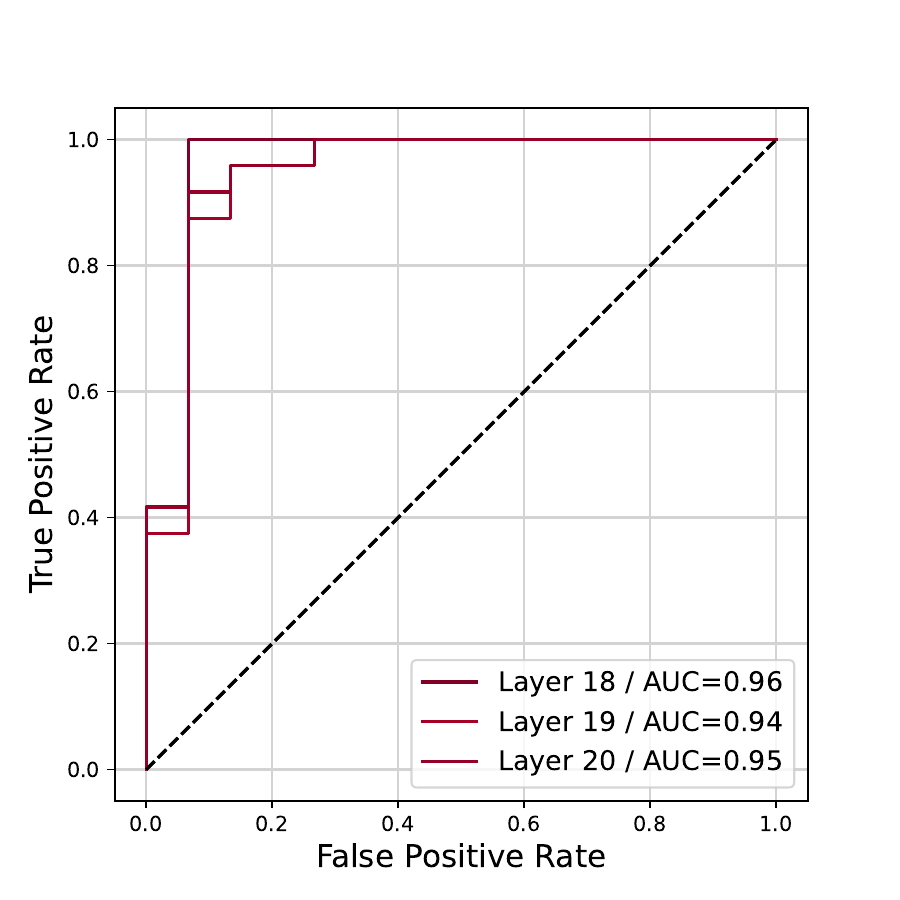}
         \caption{Corresponding activation vectors}
         \label{fig:roc_shake_acti_fair}
     \end{subfigure}
     \hfill
     \begin{subfigure}[t]{0.329\textwidth}
        \captionsetup{justification=centering}
         \centering
         \includegraphics[width=\linewidth]{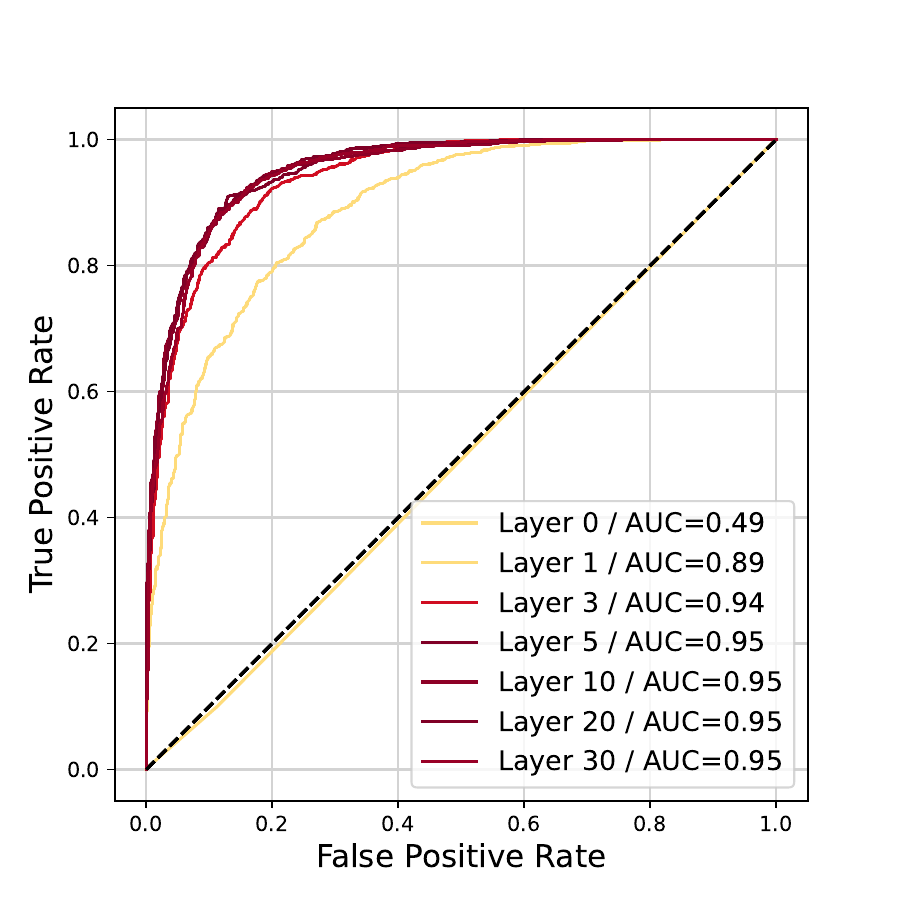}
         \caption{Activation vectors of 17k sentences}
         \label{fig:roc_shake_acti_all}
     \end{subfigure}
        \caption{Comparison between the classification results on the Shakespeare dataset: Using (a) only the trained steering vectors, (b) the corresponding activation vectors, and (c) activation vectors of 17k sentences for selected layers.}
        \label{fig:roc_curves_shake}
\end{figure*}

\paragraph{Shakespeare} The capabilities of the trained steering vectors $\mathbf{z^{(i)}_x}$ and activations $\mathbf{a^{(i)}(x)}$ at layer $i$ to encode style in the Shakespeare dataset are presented in Fig.~\ref{fig:roc_curves_shake}. In contrast to the Yelp review dataset, we want to differentiate between modern and original Shakespearean phrases. This task differs from the other two datasets in that we do not change emotion or sentiment but a whole writing style. 
The Shakespeare classifier on the trained steering vectors reaches a maximal AUC value of $0.8$, while their corresponding activation vectors reach an AUC value of $0.96$. Again, the layers $i \in \{18,19,20\}$ had high AUC values. This supports our initial findings on the Yelp review dataset.
As can be seen by comparing the AUC values for the activation vectors from Shakespeare (max. AUC = 0.96/ Fig.~\ref{fig:roc_shake_acti_all}) with Yelp in the same setting (max. AUC = 0.99/ Fig.~\ref{fig:roc_shake_acti_all}), the style difference between original and modern Shakespeare is harder to distinguish, than the sentiment in the Yelp reviews.    

\begin{figure*}[h]
     \centering
     \begin{subfigure}[t]{0.329\textwidth}
        \captionsetup{justification=centering}
         \centering
         \includegraphics[width=\linewidth]{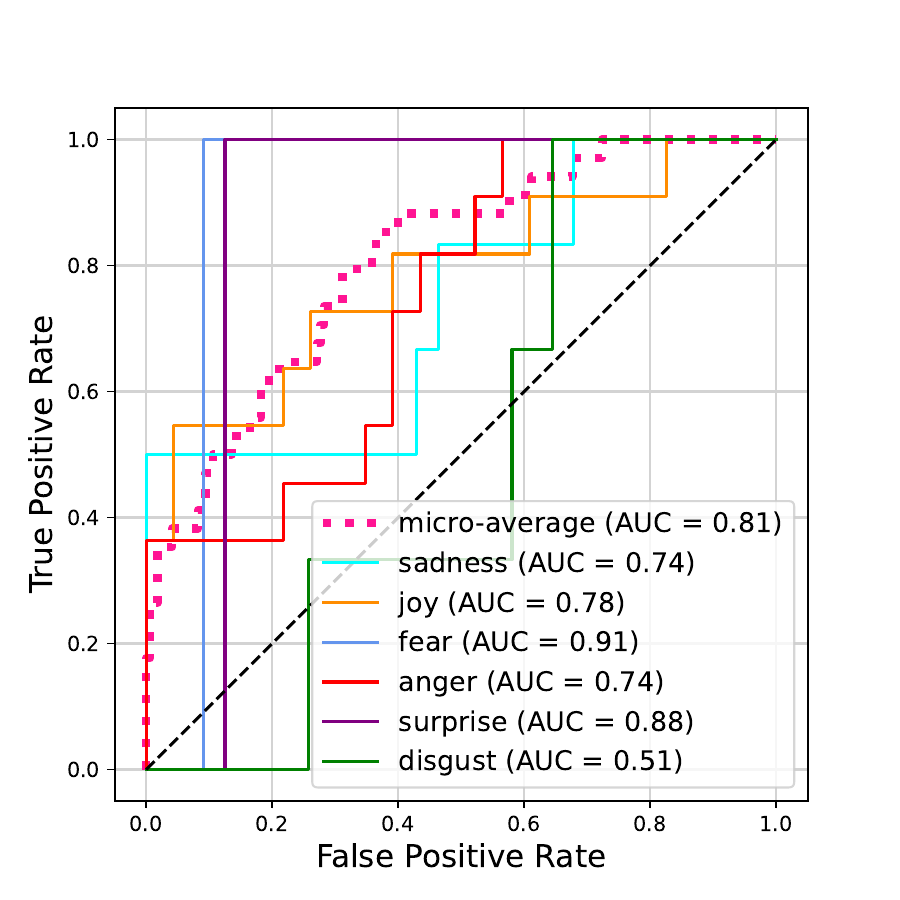}
         \caption{Trained steering vectors}
         \label{fig:roc_goemo_steering_18}
     \end{subfigure}
     \hfill
     \begin{subfigure}[t]{0.329\textwidth}
        \captionsetup{justification=centering}
         \centering
         \includegraphics[width=\linewidth]{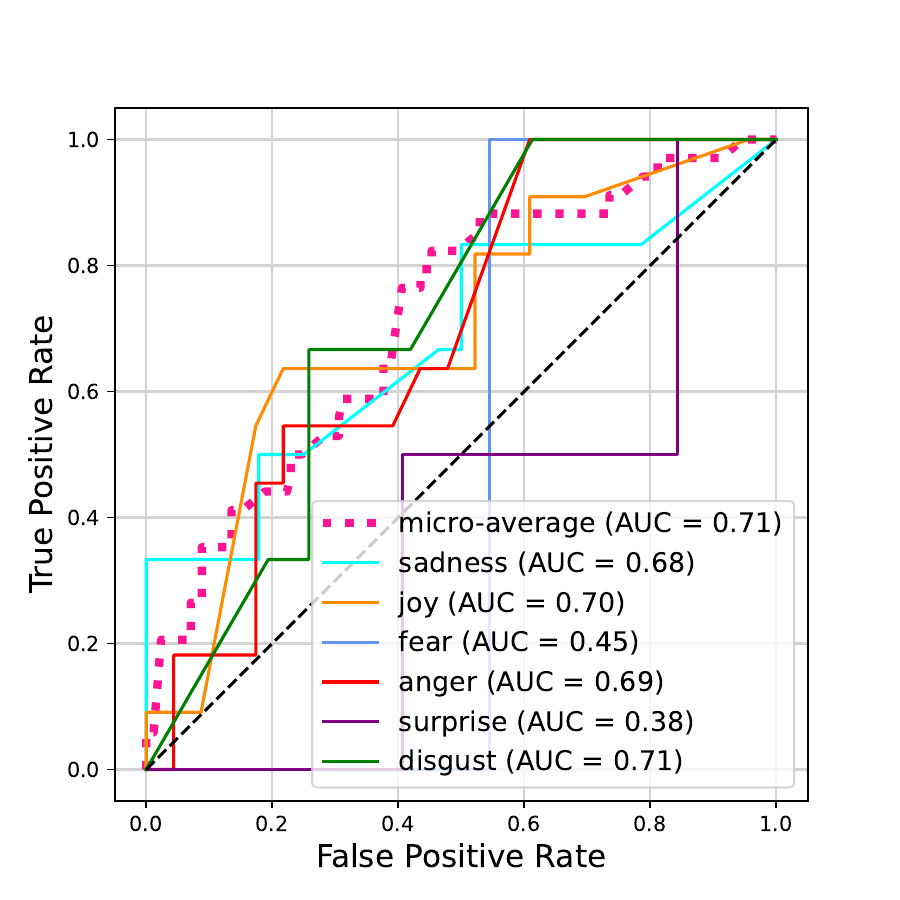}
         \caption{Corresponding activation vectors}
         \label{fig:roc_goemo_acti_fair_18}
     \end{subfigure}
     \hfill
     \begin{subfigure}[t]{0.329\textwidth}
        \captionsetup{justification=centering}
         \centering
         \includegraphics[width=\linewidth]{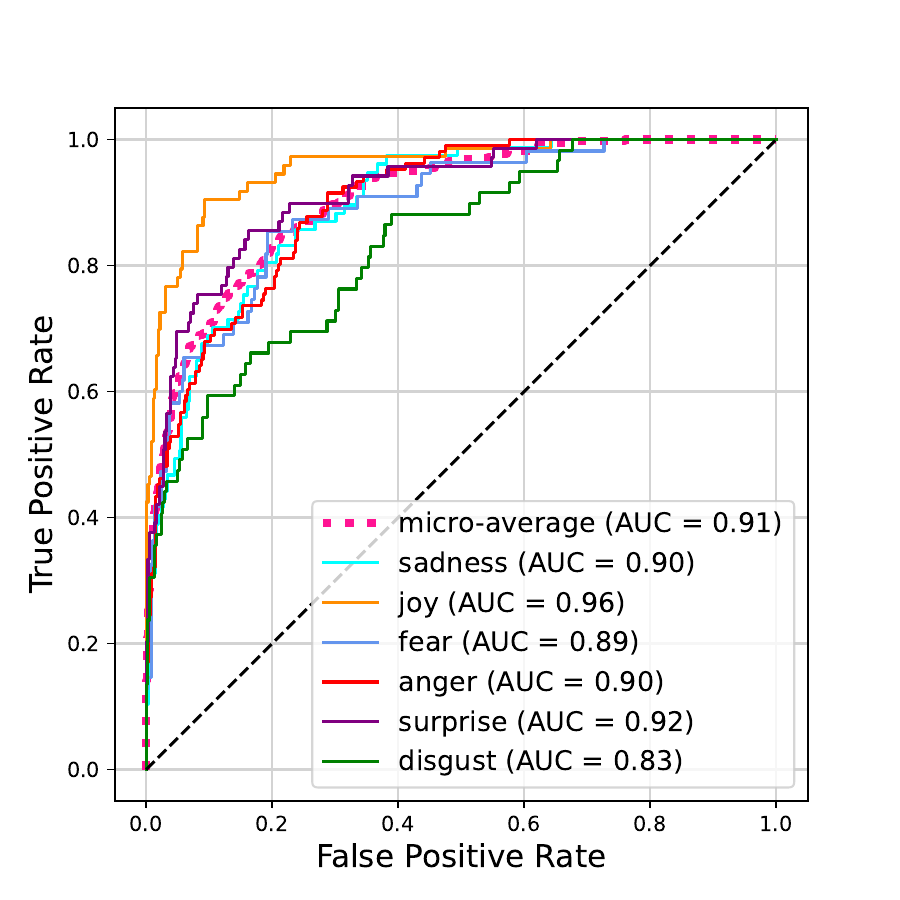}
         \caption{Activation vectors of 2k sentences}
         \label{fig:roc_goemo_acti_all_18}
     \end{subfigure}
        \caption{Classification results of vectors from layer 18 on the GoEmotions dataset: Using (a) only the trained steering vectors, (b) the corresponding activation vectors, and (c) activation vectors of 2k sentences. The activation vectors only show superior performance if we include more sentences than we have trained steering vectors.}
        \label{fig:roc_curves_goemo_18}
\end{figure*}

\begin{figure*}[h]
     \centering
     \begin{subfigure}[t]{0.329\textwidth}
        \captionsetup{justification=centering}
         \centering
         \includegraphics[width=\linewidth]{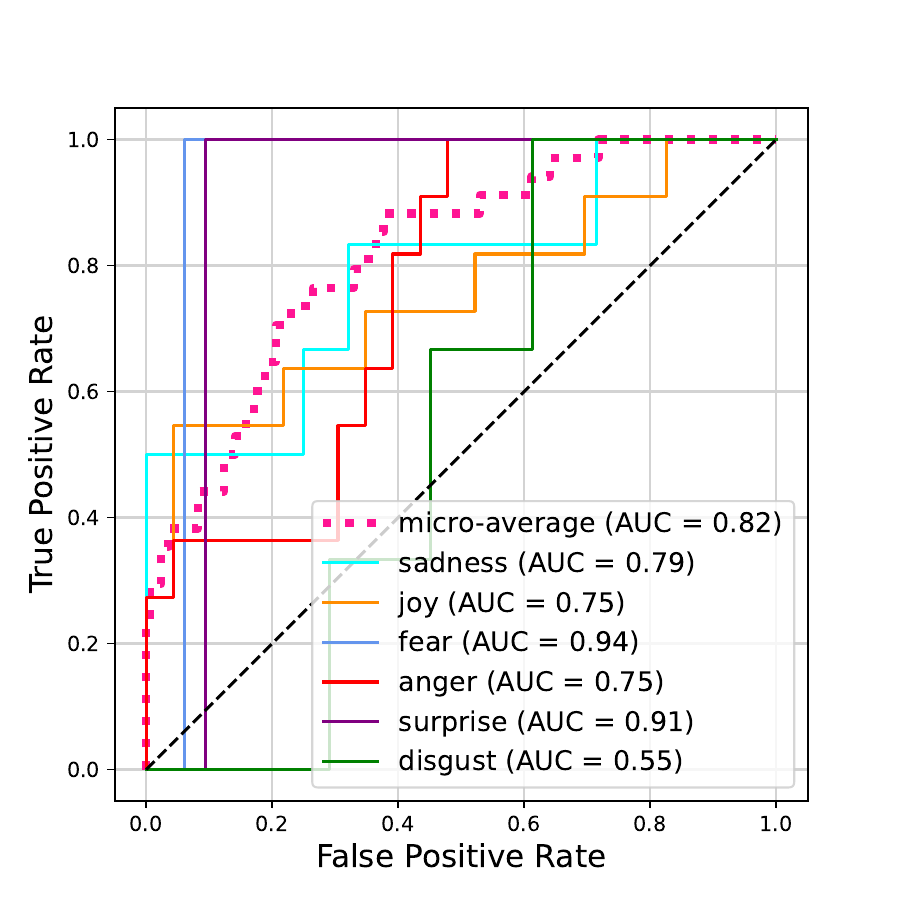}
         \caption{Trained steering vectors}
         \label{fig:roc_goemo_steering_19}
     \end{subfigure}
     \hfill
     \begin{subfigure}[t]{0.329\textwidth}
        \captionsetup{justification=centering}
         \centering
         \includegraphics[width=\linewidth]{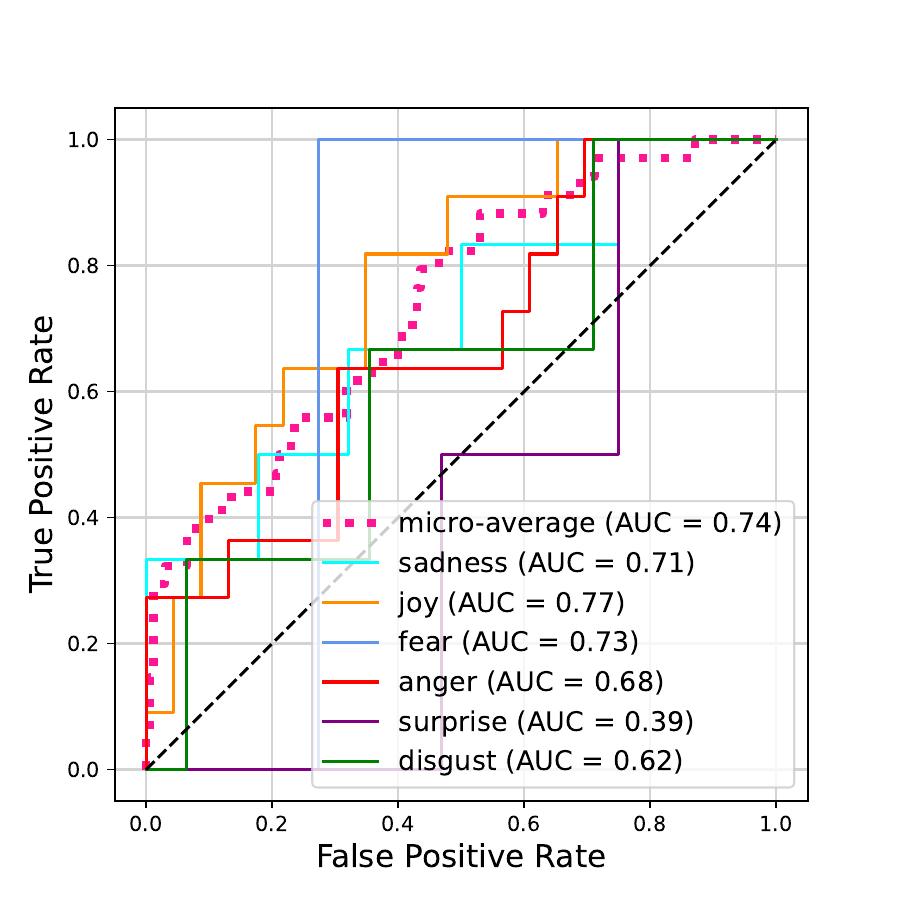}
         \caption{Corresponding activation vectors}
         \label{fig:roc_goemo_acti_fair_19}
     \end{subfigure}
     \hfill
     \begin{subfigure}[t]{0.329\textwidth}
        \captionsetup{justification=centering}
         \centering
         \includegraphics[width=\linewidth]{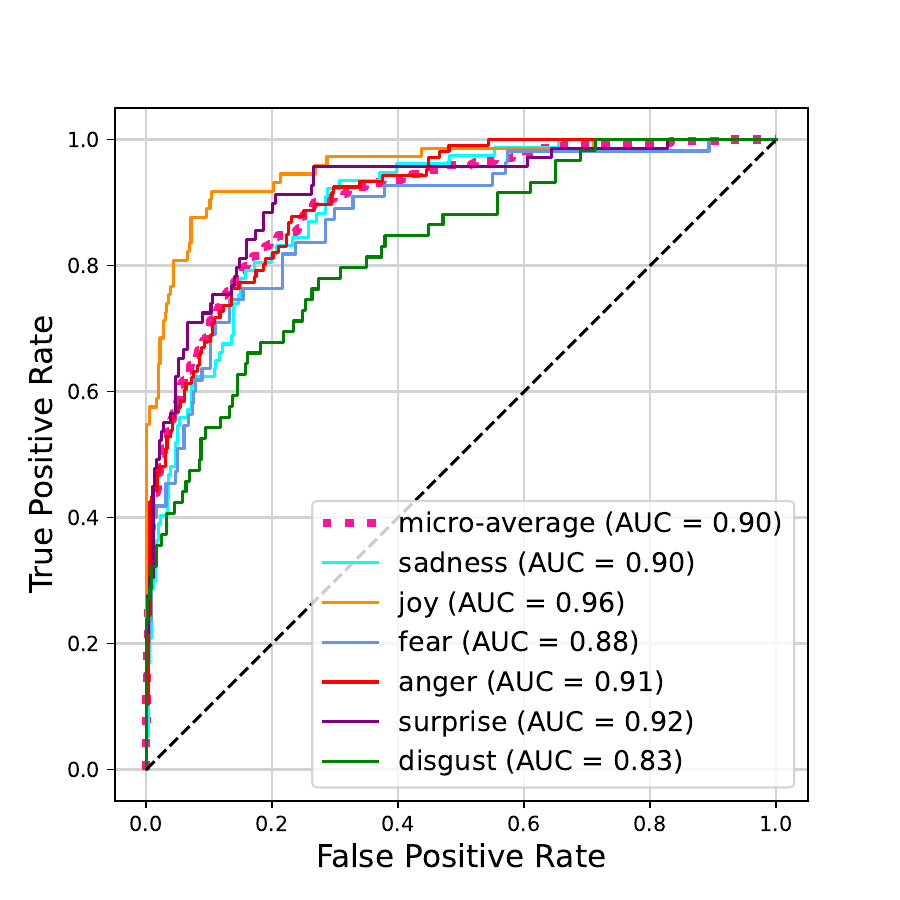}
         \caption{Activation vectors of 2k sentences}
         \label{fig:roc_goemo_acti_all_19}
     \end{subfigure}
        \caption{Classification results of vectors from layer 19 on the GoEmotions dataset: Using (a) only the trained steering vectors, (b) the corresponding activation vectors, and (c) activation vectors of 2k sentences. The activation vectors only show superior performance if we include more sentences than we have trained steering vectors.}
        \label{fig:roc_curves_goemo_19}
\end{figure*}

\begin{figure*}[h]
     \centering
     \begin{subfigure}[t]{0.329\textwidth}
        \captionsetup{justification=centering}
         \centering
         \includegraphics[width=\linewidth]{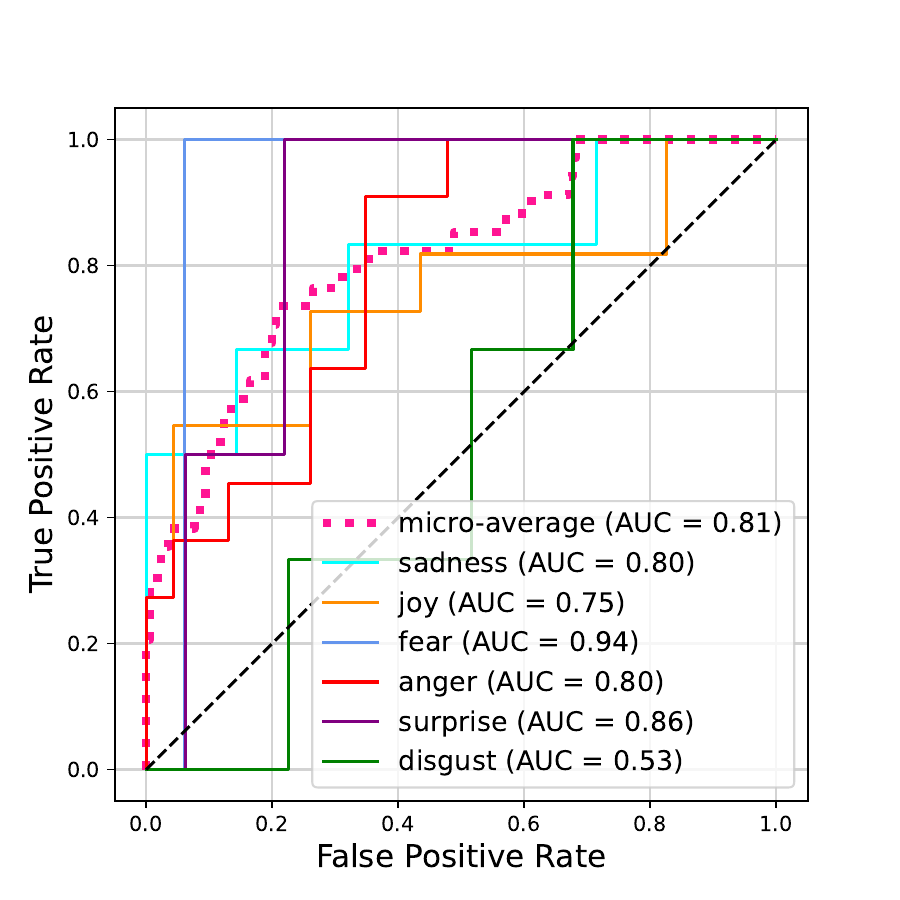}
         \caption{Trained steering vectors}
         \label{fig:roc_goemo_steering_20}
     \end{subfigure}
     \hfill
     \begin{subfigure}[t]{0.329\textwidth}
        \captionsetup{justification=centering}
         \centering
         \includegraphics[width=\linewidth]{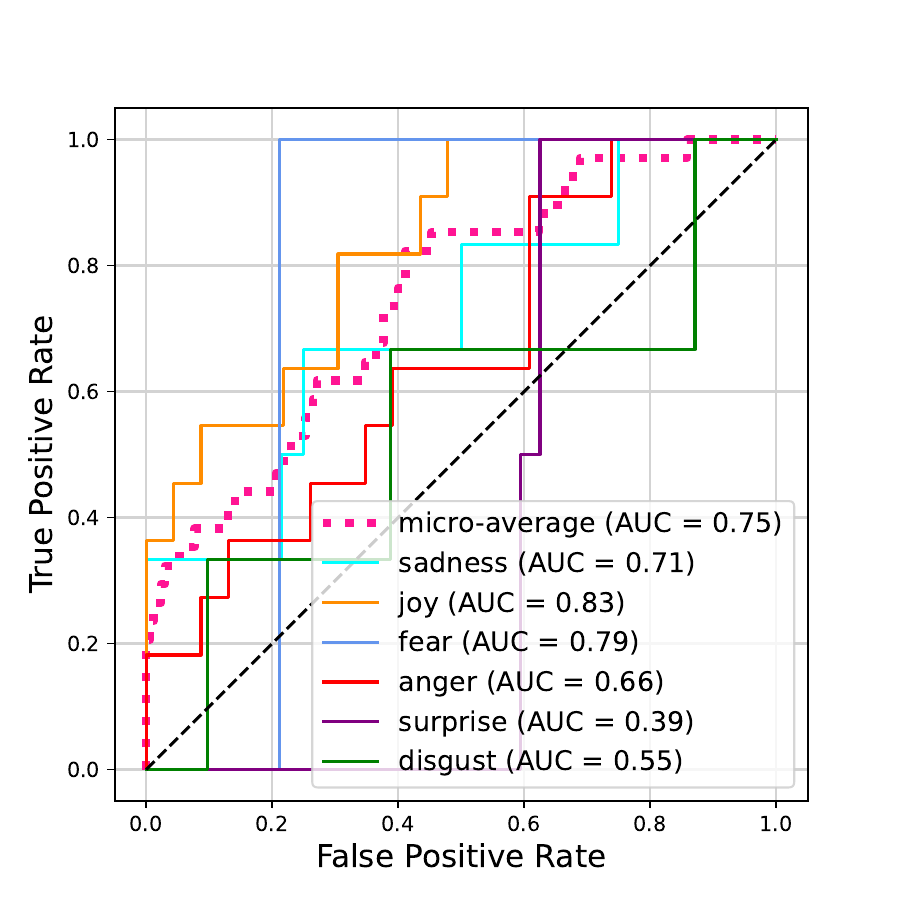}
         \caption{Corresponding activation vectors}
         \label{fig:roc_goemo_acti_fair_20}
     \end{subfigure}
     \hfill
     \begin{subfigure}[t]{0.329\textwidth}
        \captionsetup{justification=centering}
         \centering
         \includegraphics[width=\linewidth]{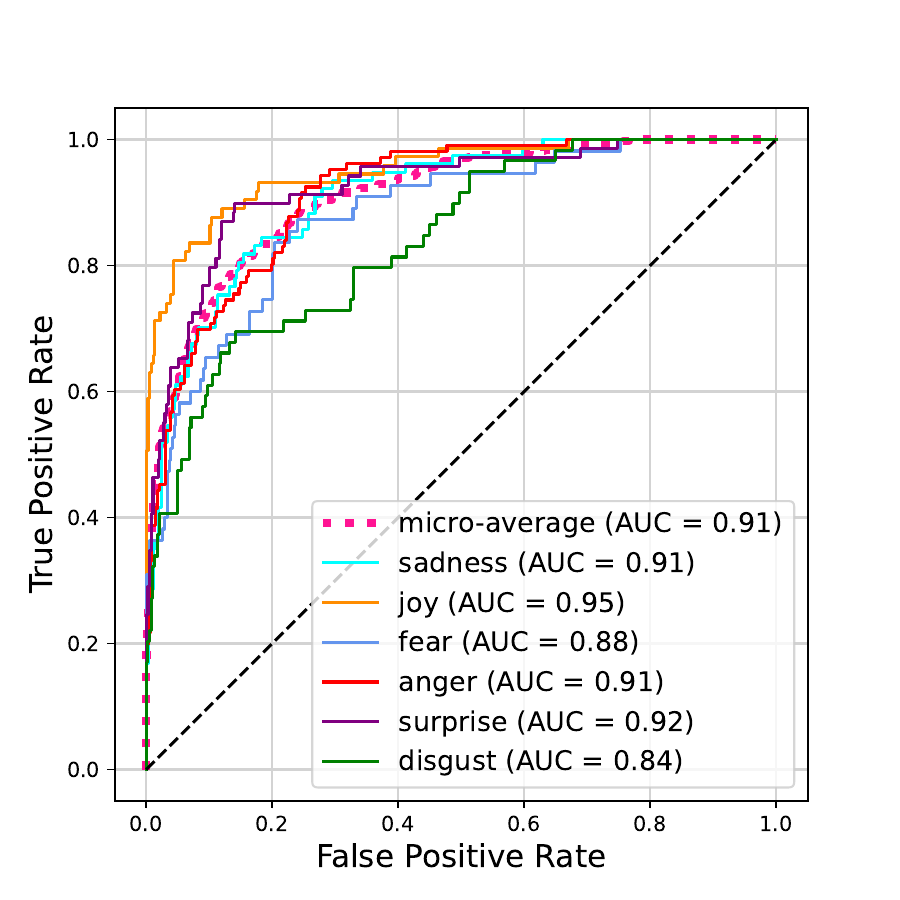}
         \caption{Activation vectors of 2k sentences}
         \label{fig:roc_goemo_acti_all_20}
     \end{subfigure}
        \caption{Classification results of vectors from layer 20 on the GoEmotions dataset: Using (a) only the trained steering vectors, (b) the corresponding activation vectors, and (c) activation vectors of 2k sentences. The activation vectors only show superior performance if we include more sentences than we have trained steering vectors.}
        \label{fig:roc_curves_goemo_20}
\end{figure*}

\paragraph{GoEmotions} For this dataset, the ROC plots need to be compared per layer because there are six instead of not two classes. The results for layer 19 draw a slightly different picture (Fig.~\ref{fig:roc_curves_goemo_19}) than for Yelp and Shakespeare. 
Probing the activations of all samples still results in the best micro-average AUC of $0.90$.
However, in the fair comparison (activations for the $89$ samples for which trained steering vectors exist), they have a micro-average AUC of $0.74$, while the corresponding trained vectors reach an AUC of $0.82$. Nevertheless, this can also result from the small number of trained steering vectors found. The same result can be seen for layers 18 (Fig.~\ref{fig:roc_curves_goemo_18}) and 20 (Fig.~\ref{fig:roc_curves_goemo_20}). We need to investigate this finding in future studies to rule out a statistical anomaly as the cause for this.  
Still, the layers $i \in \{18,19,20\}$ have high micro-average AUC values of around $0.91$ for all activations and $0.81$ for the trained steering vectors. 

\paragraph{Classifier training} During our experiments, we tried training the regression model in three different settings: Predicting the class using only a single layer, using three subsequent layers, and training on all layers together. The difference between the resulting classifications is minimal, albeit performance slightly increases when using more layers. For ease of presentation and readability of the plots, we decided to only include single-layer classifiers.

\section{Further classification-based evaluation results for output steering}
\label{S:Further_classification_based_eval}

\begin{figure*}[ht]
     \centering
     \begin{subfigure}[t]{0.329\textwidth}
        \captionsetup{justification=centering}
         \centering
         \includegraphics[width=\linewidth]{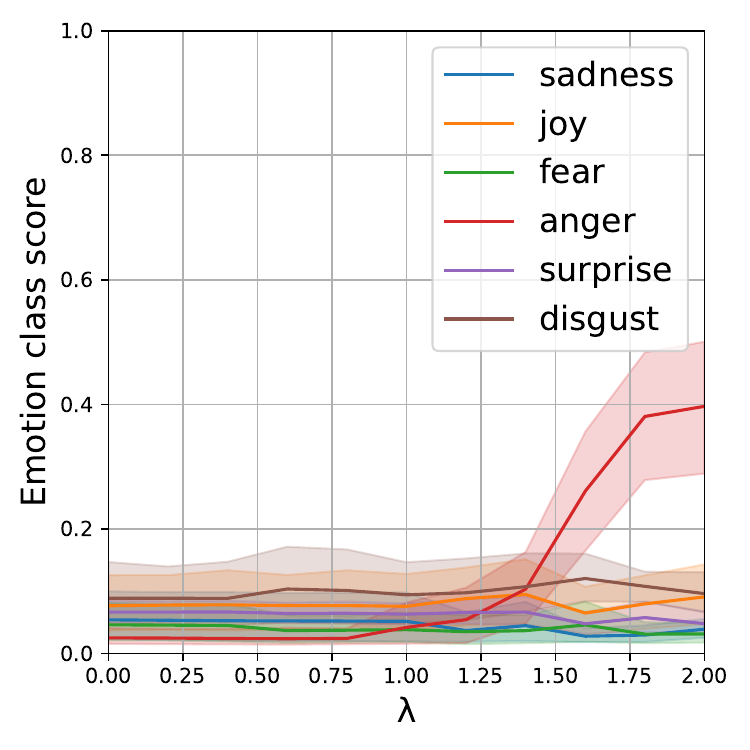}
         \caption{Steering to anger, \\factual prompts}
         \label{fig:contrastive_steering_trained_vector_based_anger_factual}
     \end{subfigure}
     \hfill
     \begin{subfigure}[t]{0.329\textwidth}
        \captionsetup{justification=centering}
         \centering
         \includegraphics[width=\linewidth]{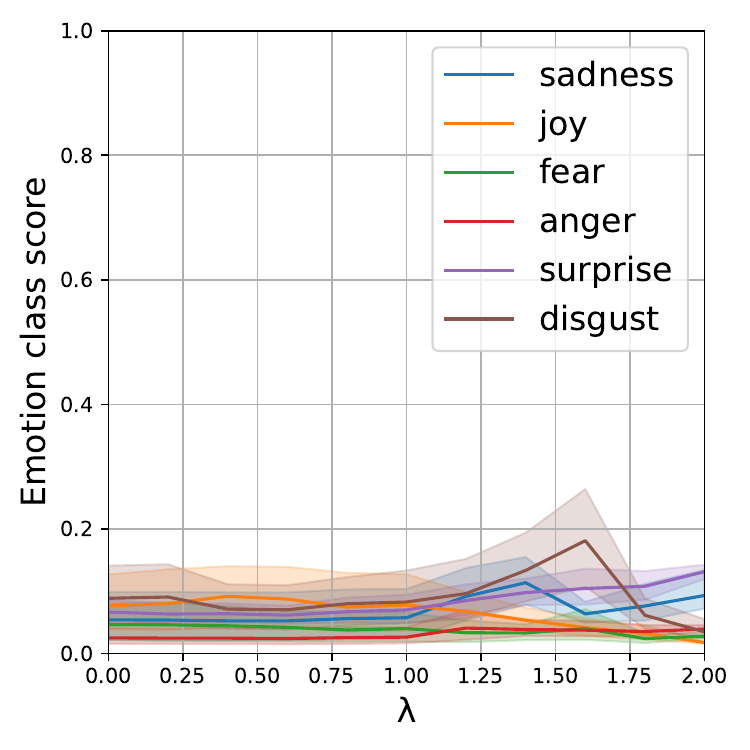}
         \caption{Steering to disgust, \\factual prompts}
         \label{fig:contrastive_steering_trained_vector_based_disgust_factual}
     \end{subfigure}
     \hfill
     \begin{subfigure}[t]{0.329\textwidth}
        \captionsetup{justification=centering}
         \centering
         \includegraphics[width=\linewidth]{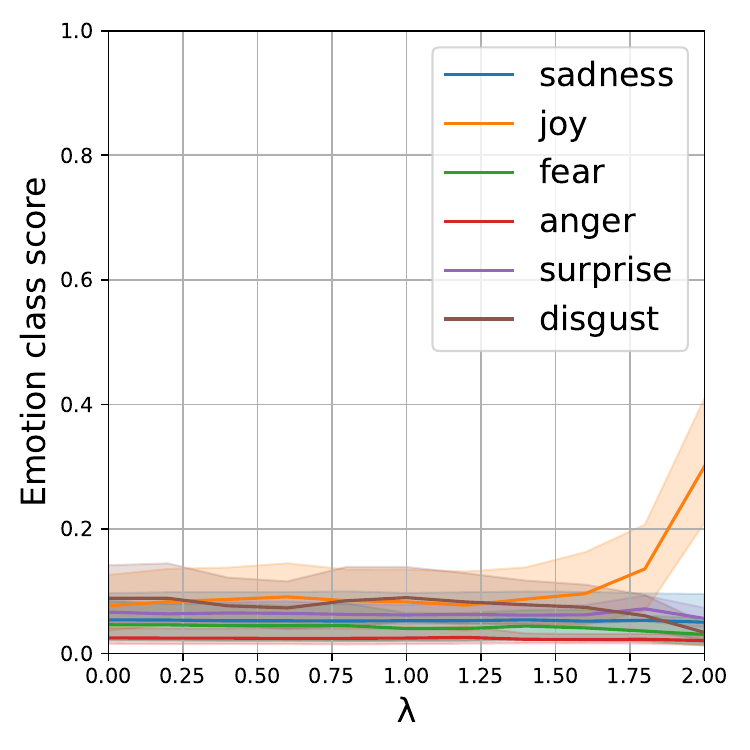}
         \caption{Steering to joy, \\factual prompts}
         \label{fig:contrastive_steering_trained_vector_based_joy_factual}
     \end{subfigure}

          \begin{subfigure}[t]{0.329\textwidth}
        \captionsetup{justification=centering}
         \centering
         \includegraphics[width=\linewidth]{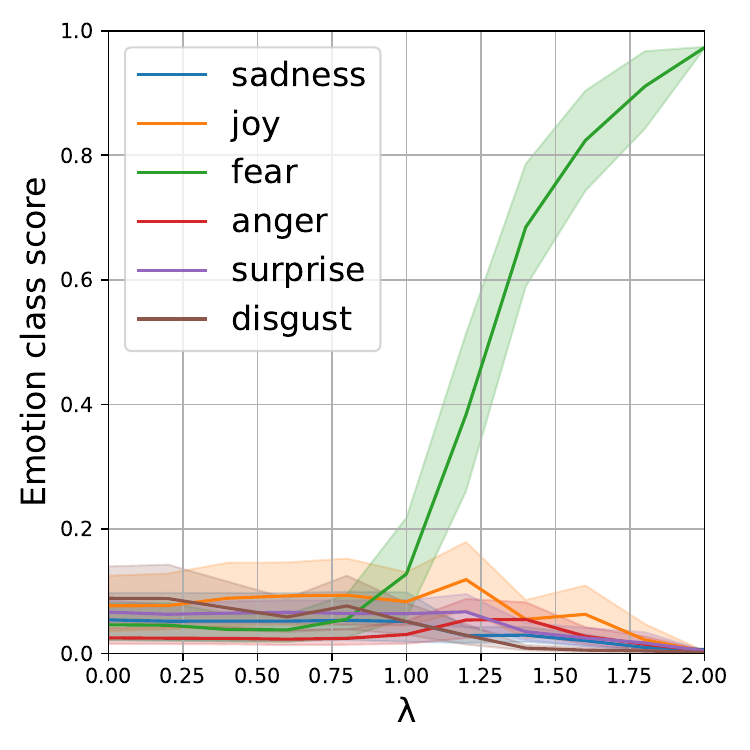}
         \caption{Steering to fear, \\factual prompts}
         \label{fig:contrastive_steering_trained_vector_based_fear_factual}
     \end{subfigure}
     \hfill
     \begin{subfigure}[t]{0.329\textwidth}
        \captionsetup{justification=centering}
         \centering
         \includegraphics[width=\linewidth]{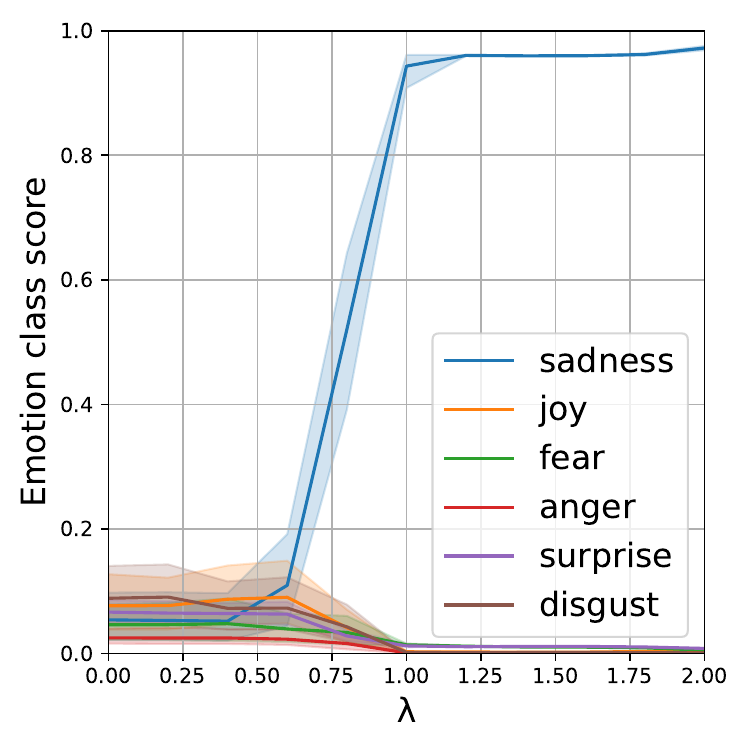}
         \caption{Steering to sadness, \\factual prompts}
         \label{fig:contrastive_steering_trained_vector_based_sadness_factual}
     \end{subfigure}
     \hfill
     \begin{subfigure}[t]{0.329\textwidth}
        \captionsetup{justification=centering}
         \centering
         \includegraphics[width=\linewidth]{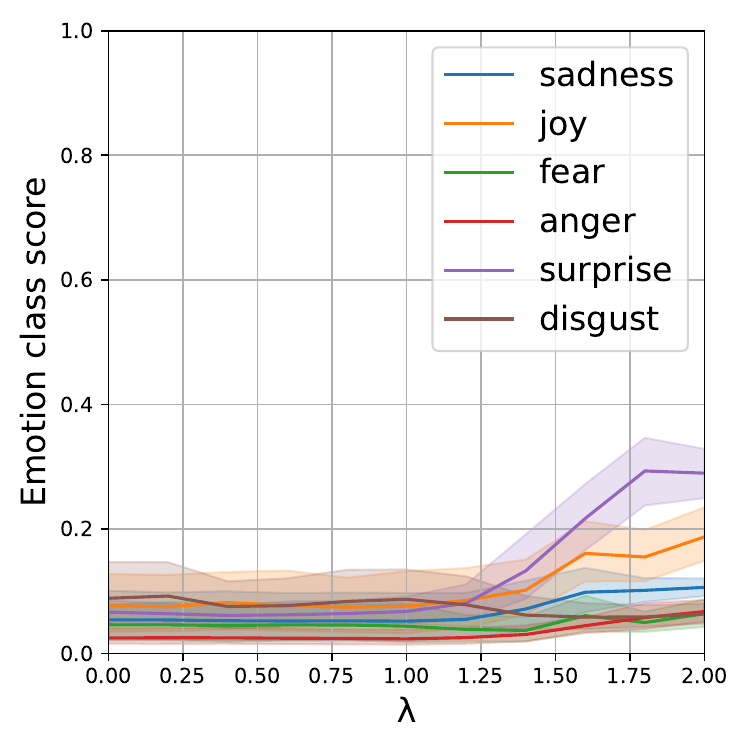}
         \caption{Steering to surprise, \\factual prompts}
         \label{fig:contrastive_steering_trained_vector_based_surprise_factual}
     \end{subfigure}
        \caption{Training-based style vectors: Evaluation of generated texts for \textit{factual} prompts using GoEmotions' style vectors.}
        \label{fig:goemo_factual_training_based}
\end{figure*}

\begin{figure*}[ht]
     \centering
     \begin{subfigure}[t]{0.329\textwidth}
        \captionsetup{justification=centering}
         \centering
         \includegraphics[width=\linewidth]{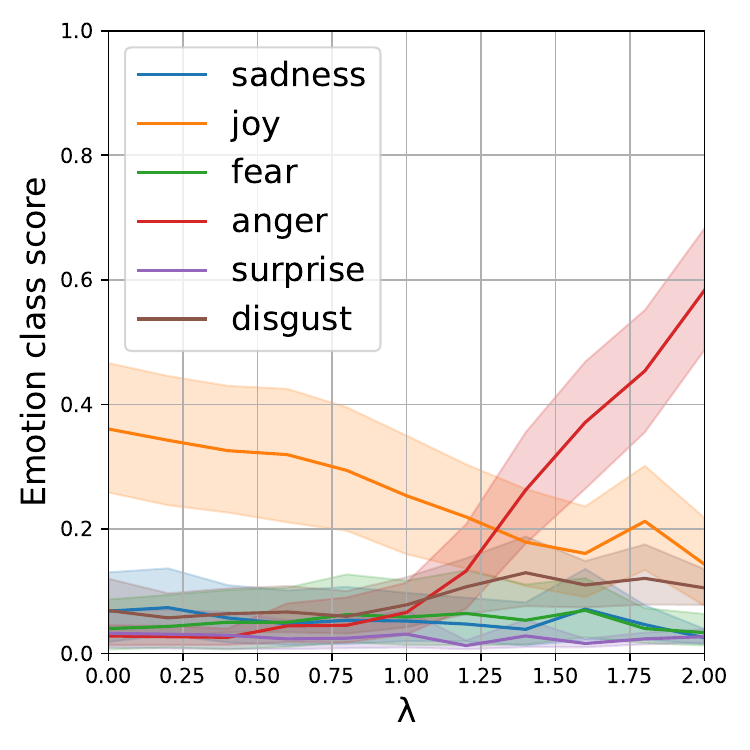}
         \caption{Steering to anger, \\subjective prompts}
         \label{fig:contrastive_steering_trained_vector_based_anger_subjective}
     \end{subfigure}
     \hfill
     \begin{subfigure}[t]{0.329\textwidth}
        \captionsetup{justification=centering}
         \centering
         \includegraphics[width=\linewidth]{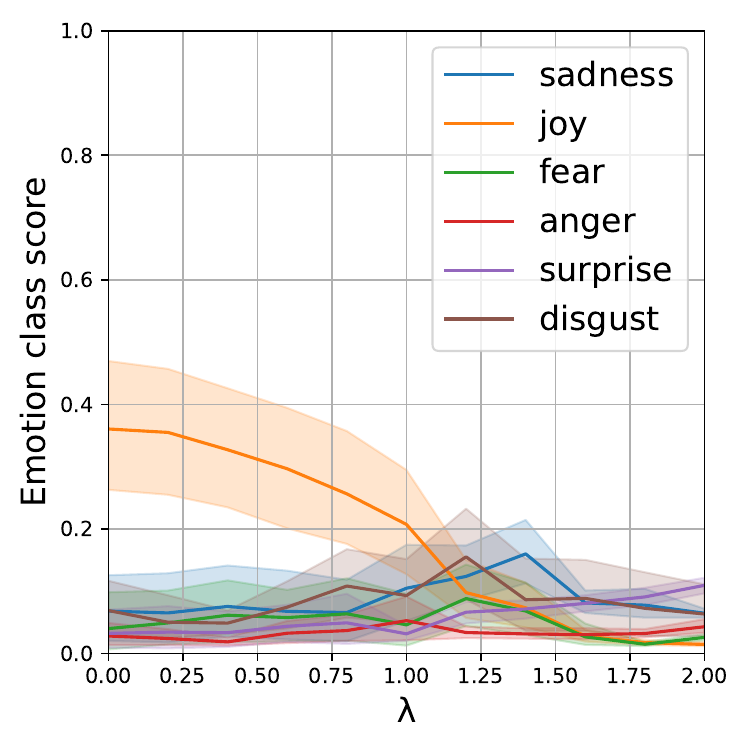}
         \caption{Steering to disgust, \\subjective prompts}
         \label{fig:contrastive_steering_trained_vector_based_disgust_subjective}
     \end{subfigure}
     \hfill
     \begin{subfigure}[t]{0.329\textwidth}
        \captionsetup{justification=centering}
         \centering
         \includegraphics[width=\linewidth]{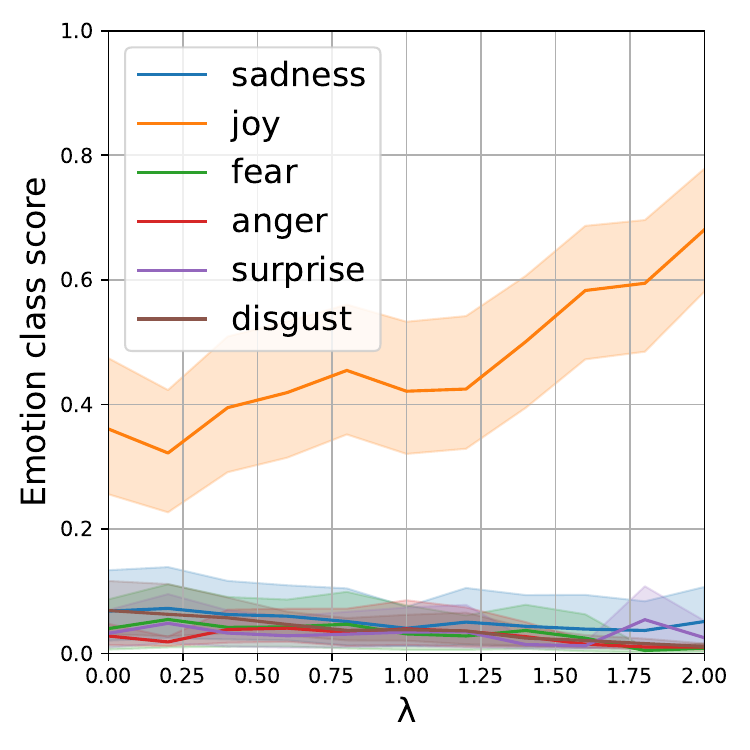}
         \caption{Steering to joy, \\subjective prompts}
         \label{fig:contrastive_steering_trained_vector_based_joy_subjective}
     \end{subfigure}

          \begin{subfigure}[t]{0.329\textwidth}
        \captionsetup{justification=centering}
         \centering
         \includegraphics[width=\linewidth]{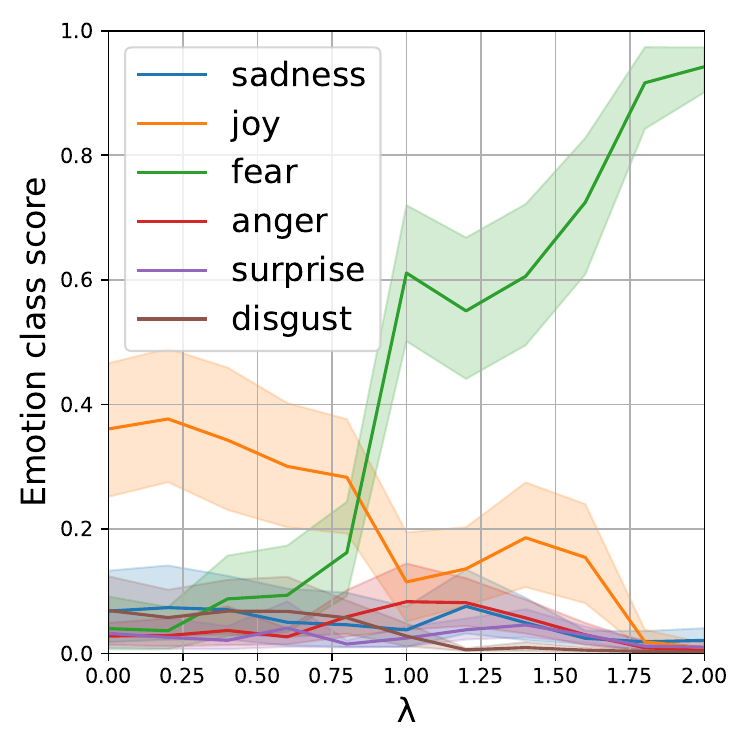}
         \caption{Steering to fear, \\subjective prompts}
         \label{fig:contrastive_steering_trained_vector_based_fear_subjective}
     \end{subfigure}
     \hfill
     \begin{subfigure}[t]{0.329\textwidth}
        \captionsetup{justification=centering}
         \centering
         \includegraphics[width=\linewidth]{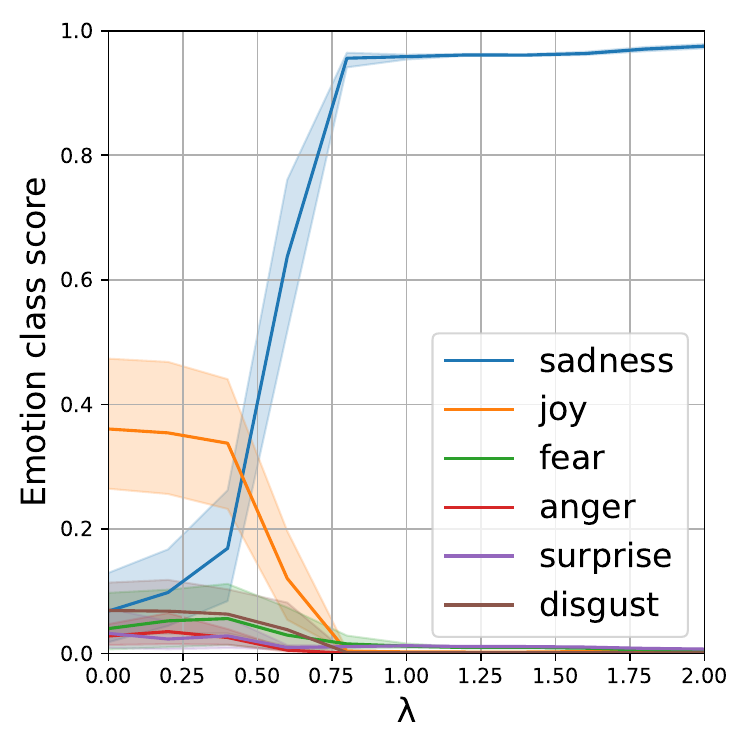}
         \caption{Steering to sadness, \\subjective prompts}
         \label{fig:contrastive_steering_trained_vector_based_sadness_subjective}
     \end{subfigure}
     \hfill
     \begin{subfigure}[t]{0.329\textwidth}
        \captionsetup{justification=centering}
         \centering
         \includegraphics[width=\linewidth]{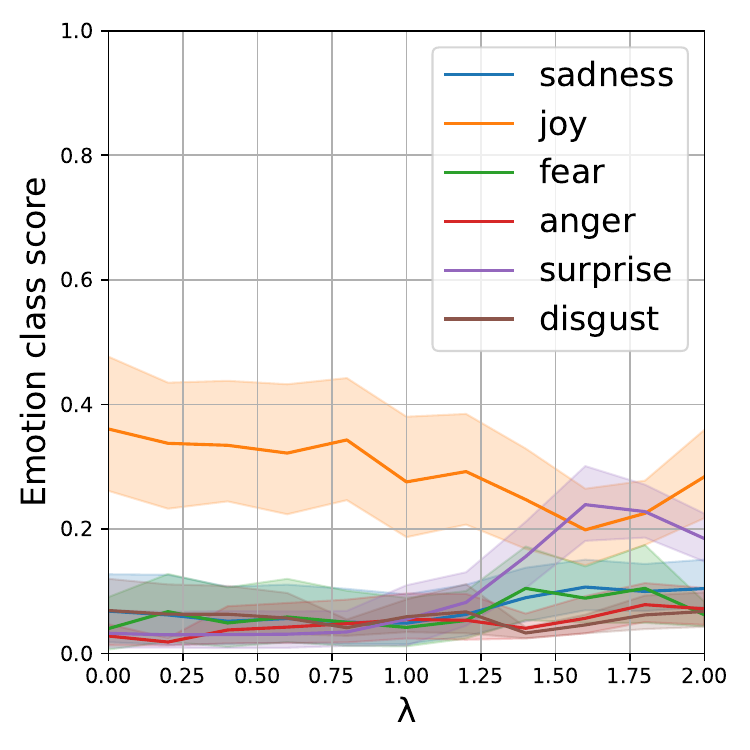}
         \caption{Steering to surprise, \\subjective prompts}
         \label{fig:contrastive_steering_trained_vector_based_surprise_subjective}
     \end{subfigure}
        \caption{Training-based style vectors: Evaluation of generated texts for \textit{subjective} prompts using GoEmotions' style vectors. Most outputs are not proper sentences.}
        \label{fig:goemo_subjective_training_based}
\end{figure*}

This section compares the training-based style vectors with their corresponding activation-based style vectors. We do this to ensure fairness in the comparison since the number of activation-based style vectors is significantly higher than the number of training-based vectors. 
In the evaluation of the factual (Fig.~\ref{fig:goemo_factual_training_based}) and subjective (Fig.~\ref{fig:goemo_subjective_training_based}) prompts using the training-based style vectors on the GoEmotions dataset, we saw that the steering seems to work for all emotions, except disgust and surprise. However, during a closer examination, it became evident that the model`s output with $\lambda \geq 0.75$ did not represent proper sentences anymore and were mainly repetitions of keywords related to the emotion, e.g., ``sadly'' for sadness. For the Yelp dataset, this happened as well, but only for higher $\lambda$. A reason for this unstable behavior in GoEmotions is probably the small number of trained steering vectors that were found, which was especially low for the classes \emph{disgust} and \emph{surprise}. 

\begin{figure*}[ht]
     \centering
     \begin{subfigure}[t]{0.329\textwidth}
        \captionsetup{justification=centering}
         \centering
         \includegraphics[width=\linewidth]{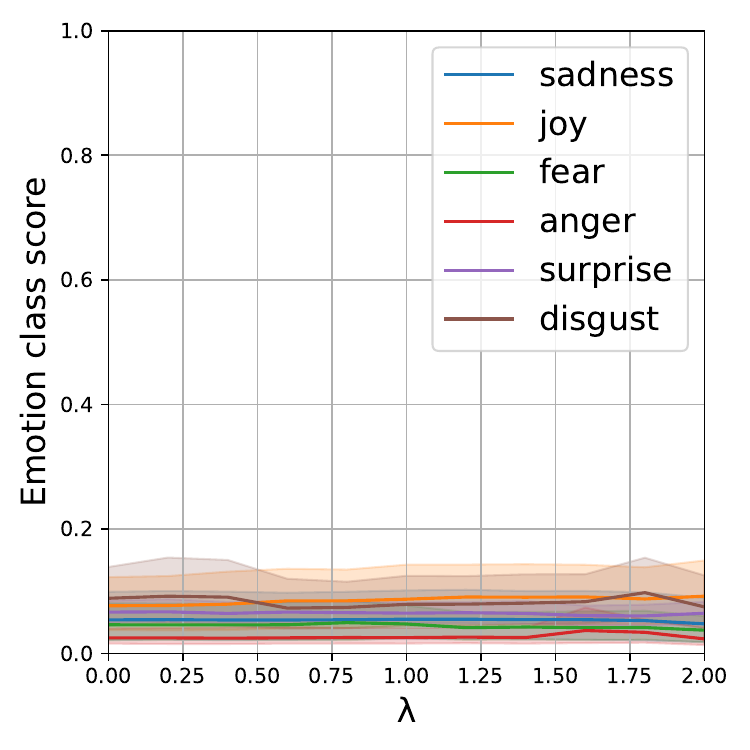}
         \caption{Steering to anger, \\factual prompts}
         \label{fig:contrastive_steering_activation_based_anger_factual}
     \end{subfigure}
     \hfill
     \begin{subfigure}[t]{0.329\textwidth}
        \captionsetup{justification=centering}
         \centering
         \includegraphics[width=\linewidth]{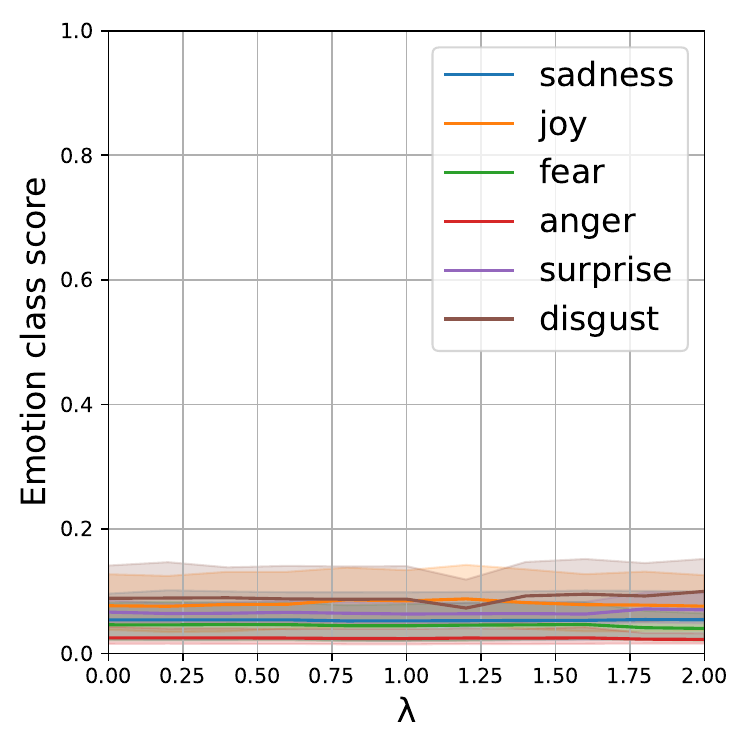}
         \caption{Steering to disgust, \\factual prompts}
         \label{fig:contrastive_steering_activation_based_disgust_factual}
     \end{subfigure}
     \hfill
     \begin{subfigure}[t]{0.329\textwidth}
        \captionsetup{justification=centering}
         \centering
         \includegraphics[width=\linewidth]{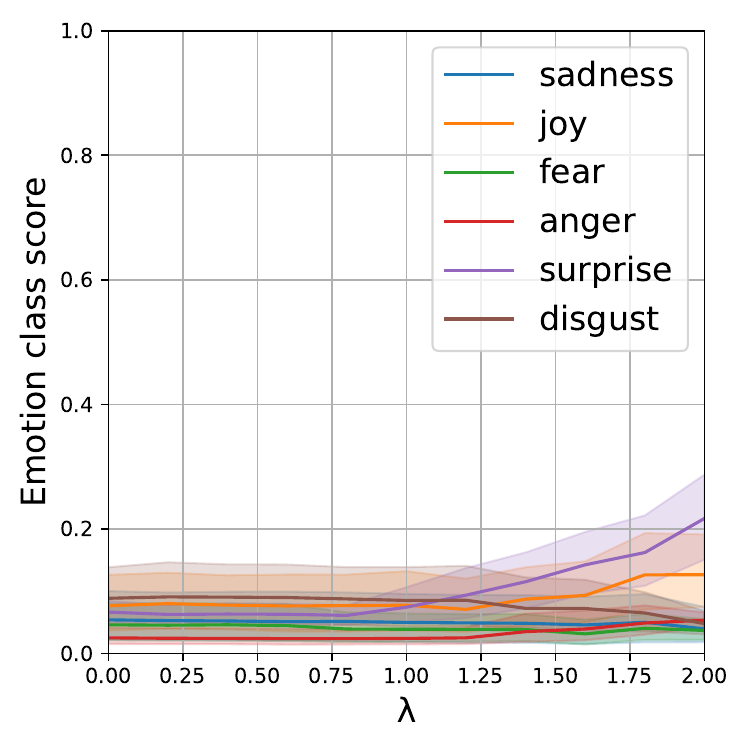}
         \caption{Steering to joy, \\factual prompts}
         \label{fig:contrastive_steering_activation_based_joy_factual}
     \end{subfigure}

          \begin{subfigure}[t]{0.329\textwidth}
        \captionsetup{justification=centering}
         \centering
         \includegraphics[width=\linewidth]{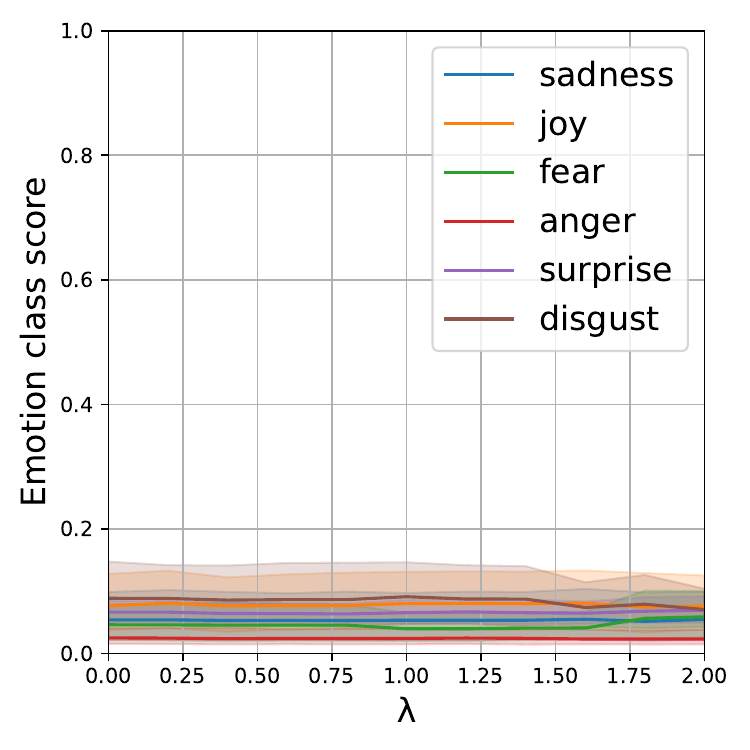}
         \caption{Steering to fear, \\factual prompts}
         \label{fig:contrastive_steering_activation_based_fear_factual}
     \end{subfigure}
     \hfill
     \begin{subfigure}[t]{0.329\textwidth}
        \captionsetup{justification=centering}
         \centering
         \includegraphics[width=\linewidth]{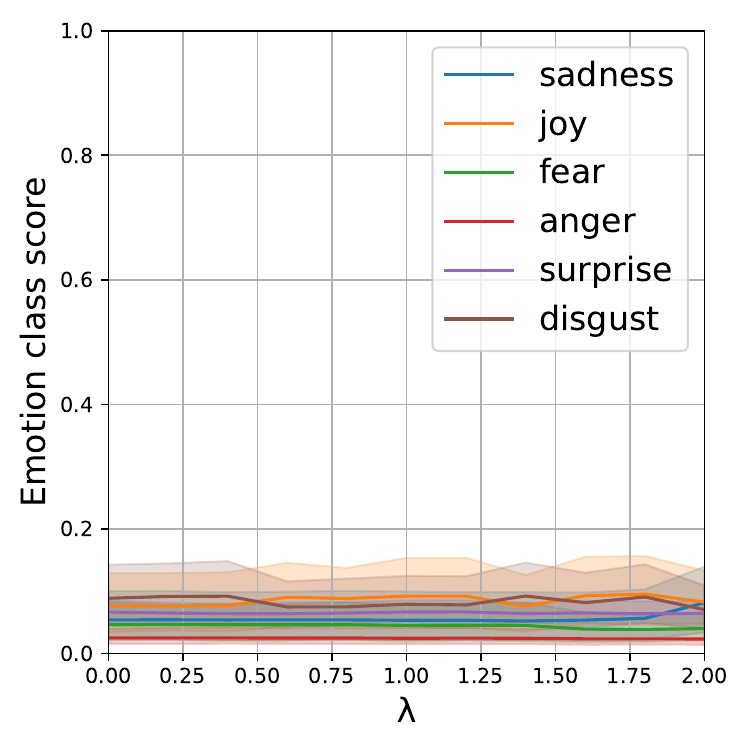}
         \caption{Steering to sadness, \\factual prompts}
         \label{fig:contrastive_steering_activation_based_sadness_factual}
     \end{subfigure}
     \hfill
     \begin{subfigure}[t]{0.329\textwidth}
        \captionsetup{justification=centering}
         \centering
         \includegraphics[width=\linewidth]{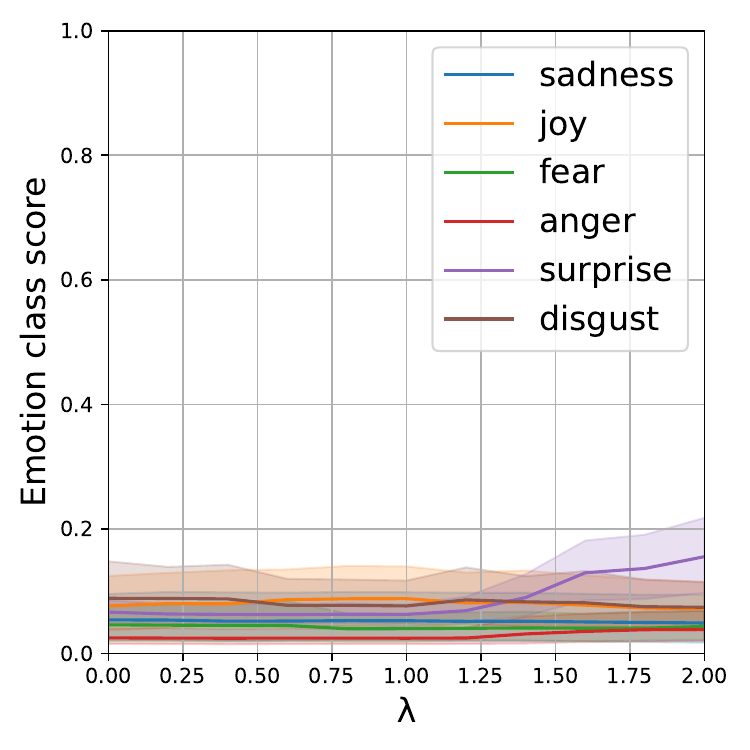}
         \caption{Steering to surprise, \\subjective prompts}
         \label{fig:contrastive_steering_activation_based_surprise_factual}
     \end{subfigure}
        \caption{Activation-based style vectors: Evaluation of generated texts for \textit{factual} prompts using GoEmotions' style vectors. Only the activation vectors were used, for which we have trained steering vectors.}
        \label{fig:goemo_factual_activations}
\end{figure*}

\begin{figure*}[ht]
     \centering
     \begin{subfigure}[t]{0.329\textwidth}
        \captionsetup{justification=centering}
         \centering
         \includegraphics[width=\linewidth]{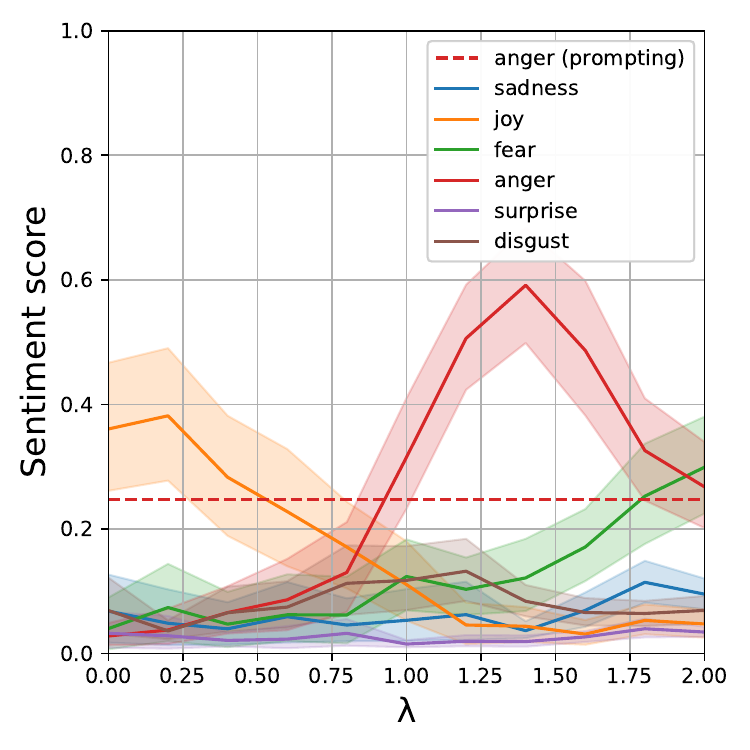}
         \caption{Steering to anger, \\subjective prompts}
         \label{fig:contrastive_steering_anger_subjective}
     \end{subfigure}
     \hfill
     \begin{subfigure}[t]{0.329\textwidth}
        \captionsetup{justification=centering}
         \centering
         \includegraphics[width=\linewidth]{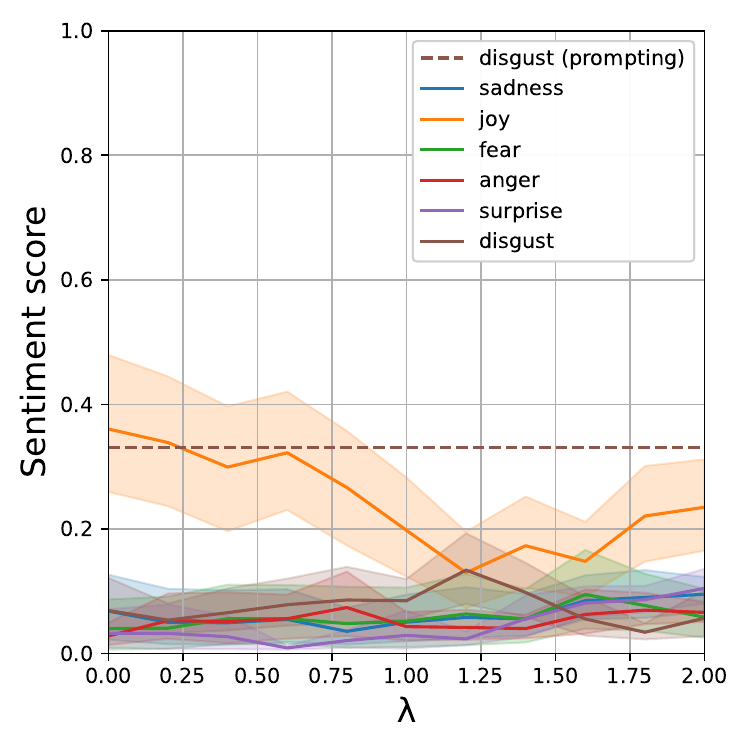}
         \caption{Steering to disgust, \\subjective prompts}
         \label{fig:contrastive_steering_disgust_subjective}
     \end{subfigure}
     \hfill
     \begin{subfigure}[t]{0.329\textwidth}
        \captionsetup{justification=centering}
         \centering
         \includegraphics[width=\linewidth]{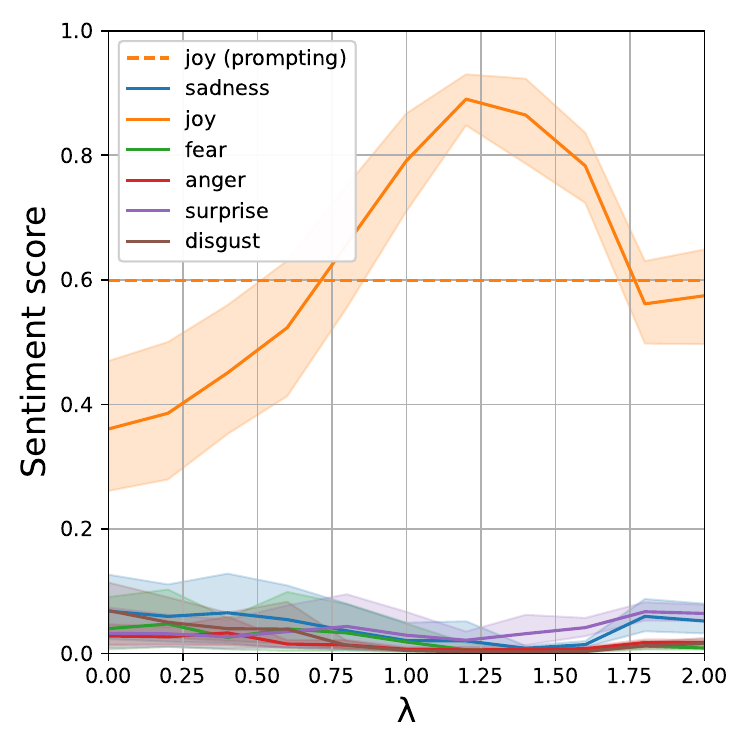}
         \caption{Steering to joy, \\subjective prompts}
         \label{fig:contrastive_steering_joy_subjective}
     \end{subfigure}

          \begin{subfigure}[t]{0.329\textwidth}
        \captionsetup{justification=centering}
         \centering
         \includegraphics[width=\linewidth]{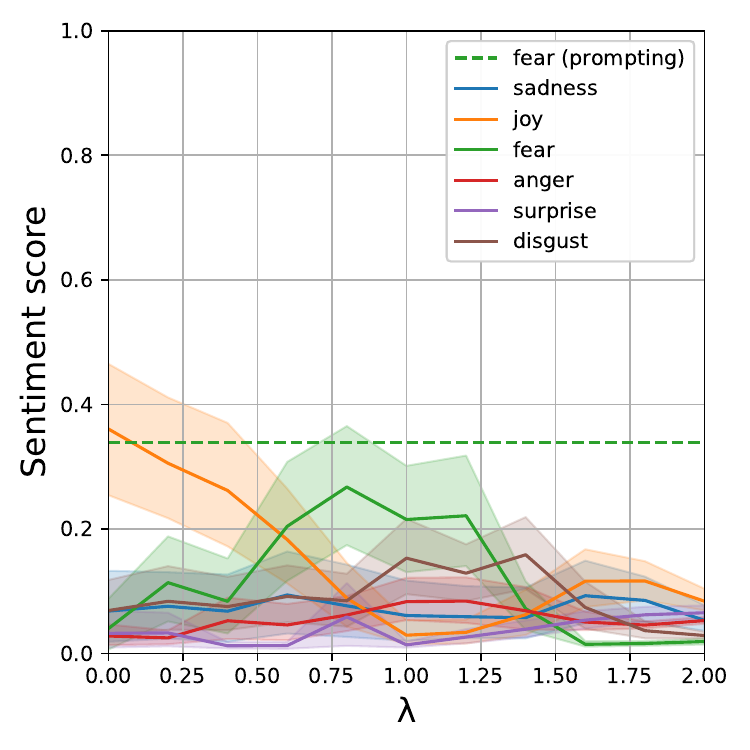}
         \caption{Steering to fear, \\subjective prompts}
         \label{fig:contrastive_steering_fear_subjective}
     \end{subfigure}
     \hfill
     \begin{subfigure}[t]{0.329\textwidth}
        \captionsetup{justification=centering}
         \centering
         \includegraphics[width=\linewidth]{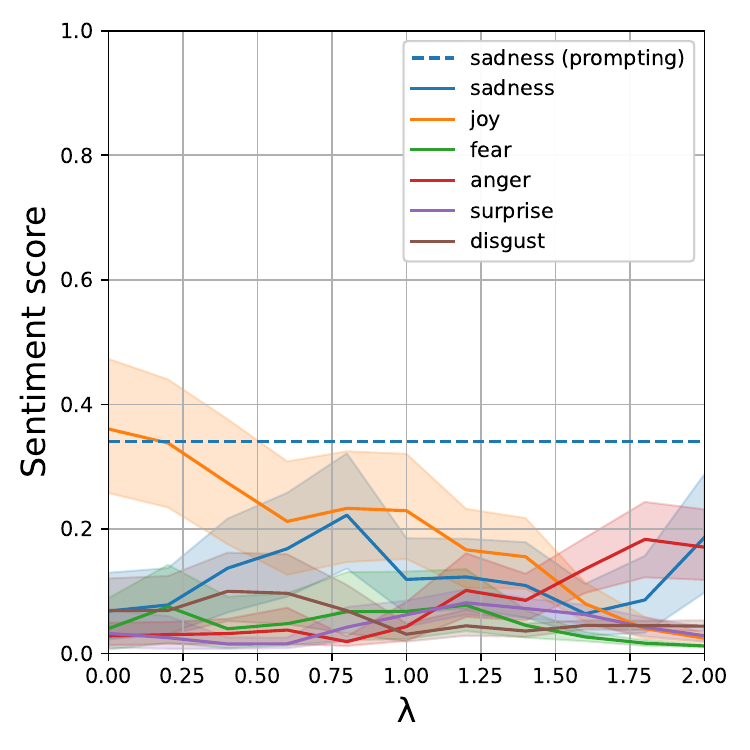}
         \caption{Steering to sadness, \\subjective prompts}
         \label{fig:contrastive_steering_sadness_subjective}
     \end{subfigure}
     \hfill
     \begin{subfigure}[t]{0.329\textwidth}
        \captionsetup{justification=centering}
         \centering
         \includegraphics[width=\linewidth]{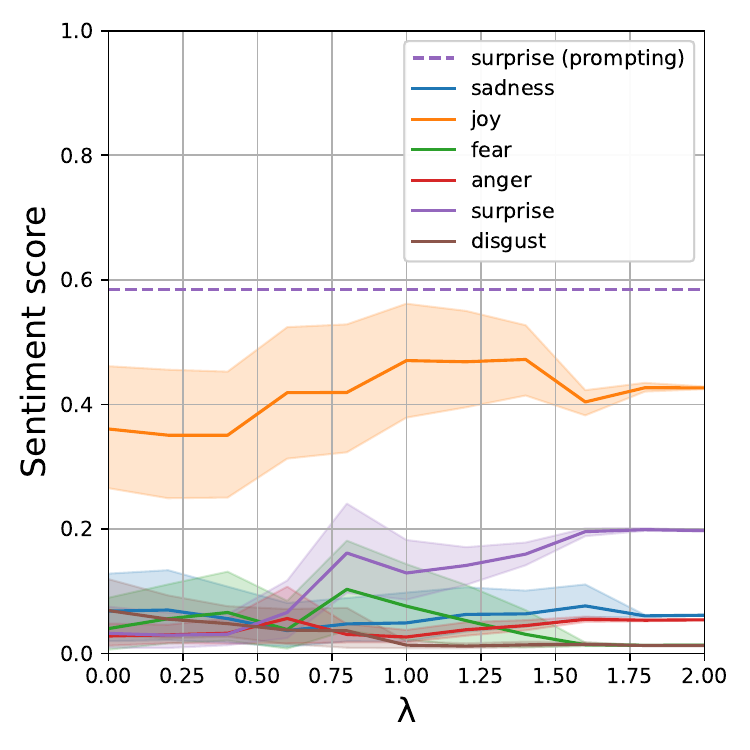}
         \caption{Steering to surprise, \\subjective prompts}
         \label{fig:contrastive_steering_surprise_subjective}
     \end{subfigure}
        \caption{Activation-based style vectors: Evaluation of generated texts for \textit{subjective} prompts using GoEmotions' style vectors. Only the activation vectors were used, for which we have trained steering vectors.} 
        \label{fig:goemo_subjective_activations}
\end{figure*}

The steering is much more stable for the activation-based style vectors for factual prompts (Fig.~\ref{fig:goemo_factual_activations}), while the subjective are not steered well (Fig.~\ref{fig:goemo_subjective_activations}) prompts. The generated sentences seem to be biased towards \emph{joy}. 
Especially, \emph{disgust} does not seem to be steered. These results, especially in comparison to the steering with all activation-based style vectors~(\ref{fig:goemo_subjective}), are, again, the result of the small number of trained steering vectors, which limits the amount of available activation-based style vectors. 
This, furthermore, highlights the superiority of the activation-based style vectors, which can be just extracted and do not require a computationally expensive learning procedure. 

\end{document}